\documentclass[sigconf]{acmart}

\usepackage{lineno,hyperref}
\usepackage{bbm}
\usepackage{bm}
\usepackage{amsfonts}
\usepackage{amsmath}
\usepackage{mathrsfs}
\usepackage{graphicx}
\usepackage{caption}
\usepackage{subfigure}
\usepackage{booktabs}
\usepackage{setspace}
\usepackage{tabularx}
\usepackage{makecell}
\usepackage{multirow}
\usepackage{bbding}
\usepackage{soul}
\usepackage{euscript}

\usepackage{multicol}

\usepackage{multirow}
\usepackage{colortbl}

\usepackage{enumitem}
\usepackage{tabularx}
\usepackage{threeparttable}
\usepackage{multirow}
\usepackage{xcolor}
\usepackage{colortbl}
\usepackage{graphicx}
\usepackage{subcaption}

\usepackage{pifont}
\newcommand{\cmark}{\ding{51}}
\newcommand{\xmark}{\ding{55}}

\usepackage{algorithm}
\usepackage{algorithmic}
\usepackage{bibunits}

\AtBeginDocument{%
	\providecommand\BibTeX{{%
			\normalfont B\kern-0.5em{\scshape i\kern-0.25em b}\kern-0.8em\TeX}}}

\copyrightyear{2026}
\acmYear{2026}
\setcopyright{acmlicensed}\acmConference[MM '26]{Proceedings of the 34rd ACM International Conference on Multimedia}{November 10--14, 2026}{Rio de Janeiro, Brazi}
\acmBooktitle{Proceedings of the 34rd ACM International Conference on Multimedia (MM '26), November 10--14, 2026, Rio de Janeiro, Brazi}
\acmDOI{xxxxxxxx.xxxxxxx}
\acmISBN{xxxxxxx.xxxxxxx}
\begin{document}
	
	\title{QuReC: All-in-One Image Restoration with Query-Specific Guidance and Local-Global Response Calibration}
	
	\author{Shen Zhou}
	\orcid{0009-0008-7119-4830}
	\affiliation{%
		\institution{Southeast University}
		\city{Nanjing}
		\country{China}
	}
	\email{zhoushen@seu.edu.cn}

	\author{Jinghui Zhang}
	\authornote{Corresponding author.}
	\orcid{0000-0002-9067-7896}
	\affiliation{%
		\institution{Southeast University}
		\city{Nanjing}
		\country{China}
	}
	\email{jhzhang@seu.edu.cn}

	\author{Wenbo Huang}
	\orcid{0000-0002-6664-1172}
	\affiliation{%
		\institution{Southeast University}
		\city{Nanjing}
		\country{China}
	}
	\email{wenbohuang1002@outlook.com}
	
	\author{Xuwei Qian}
	\orcid{0000-0002-3321-8327}
	\affiliation{%
		\institution{Southeast University}
		\city{Nanjing}
		\country{China}
	}
	\email{xuwei.qian@seu.edu.cn}
	
	\author{Zhen Wu}
	\orcid{0009-0003-6221-280X}
	\affiliation{%
		\institution{Southeast University}
		\city{Nanjing}
		\country{China}
	}
	\email{zhen-wu@seu.edu.cn}
	
	\author{Guangwen Peng}
	\orcid{0009-0008-1434-7319}
	\affiliation{%
		\institution{Hainan University}
		\city{Hainan}
		\country{China}
	}
	\email{gwen@hainanu.edu.cn}
	
	\author{Zhiyuan Li}
	\orcid{0009-0009-2804-376X}
	\affiliation{%
		\institution{Hainan University}
		\city{Hainan}
		\country{China}
	}
	\email{lizhiyuan@hainanu.edu.cn}
	
	\author{Ding Ding}
	\orcid{0000-0001-6597-3725}
	\affiliation{%
		\institution{Southeast University}
		\city{Nanjing}
		\country{China}
	}
	\email{dingding-1@seu.edu.cn}
	
	\author{Dian Shen}
	\orcid{0000-0003-0422-5285}
	\affiliation{%
		\institution{Southeast University}
		\city{Nanjing}
		\country{China}
	}
	\email{dshen@seu.edu.cn}
	
	\author{Fang Dong}
	\orcid{0000-0001-6770-326X}
	\affiliation{%
		\institution{Southeast University}
		\city{Nanjing}
		\country{China}
	}
	\email{fdong@seu.edu.cn}
	
	\renewcommand{\shortauthors}{Shen Zhou, et al.}

	\begin{abstract}
		All-in-one image restoration aims to recover clean images degraded by multiple corruption types using a single unified model. Existing methods typically rely on image-level prompts or shared guidance to handle diverse degradations. However, such a paradigm becomes inadequate when degradations are spatially heterogeneous or even coexist in mixed forms within a single image. Yet spatially adaptive guidance alone is not sufficient, since accurate restoration also requires each spatial query to reliably aggregate complementary information from local neighborhoods and global contexts. To this end, we propose QuReC, a unified framework for all-in-one image restoration. QuReC consists of a Degradation-Guided Query Reconstruction Module (DQRM) and a Local-Global Response Calibration Module (LGRCM). Specifically, DQRM matches each spatial query against a degradation prototype space to reconstruct a query-specific degradation-aware representation, thereby providing fine-grained spatially adaptive restoration guidance. To further stabilize this query-wise matching process, we introduce a weakly supervised prototype matching learning strategy to improve optimization stability and degradation semantic consistency. Meanwhile, LGRCM performs local-global dual-branch aggregation and calibrates the aggregated responses with learnable priors, improving the reliability of feature aggregation and the coordination between local detail modeling and global context modeling. Extensive experiments demonstrate that QuReC achieves superior performance on multiple all-in-one image restoration benchmarks. The code is released at \textcolor{purple}{\url{https://github.com/zhoushen1/QuReC}}.
	\end{abstract}

	
	
	\maketitle

\section{Introduction}
Image restoration is a classical inverse problem in computer vision, aiming to recover high-quality clean images from observations degraded by haze, rain streaks, noise, blur, low illumination, and other degradation factors. Owing to its importance for many downstream vision tasks \cite{1,2,3,4}, image restoration has long attracted extensive attention. In recent years, significant progress has been made in restoration methods tailored to specific degradation types, such as image denoising \cite{5,6,7,8,9}, deblurring \cite{10,11,12,13,14}, deraining \cite{15,16,17,19}, dehazing \cite{20,21,22,23,24}, and low-light enhancement \cite{25,26,27,28,29}. These methods usually achieve excellent performance on their respective tasks, but most of them are designed for only a single degradation type. When the degradation category, degradation severity, or spatial distribution varies, or when multiple degradations coexist in one image, these single-task methods often fail to maintain stable and satisfactory restoration performance. These limitations have motivated the development of all-in-one image restoration models, which aim to handle diverse restoration tasks within a unified framework \cite{30,31}. Despite encouraging progress, how to effectively adapt a single model to diverse degradations while preserving restoration quality across different tasks remains a challenging problem.

\begin{figure}[tb]
	\centering
	\includegraphics[width=1\linewidth]{{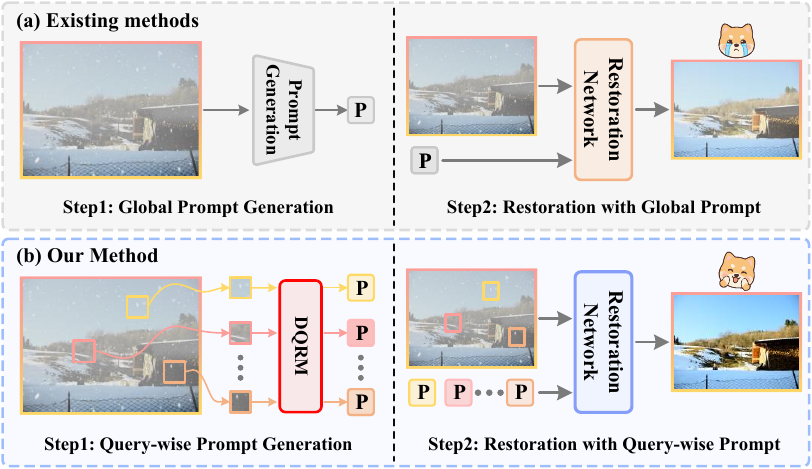}}  
	\vspace{-6mm}  
	\caption{Comparison of existing prompt-based methods and our method. (a) Existing methods use a global prompt for image-level guidance. (b) Our method uses query-specific guidance for spatially adaptive restoration.}
	\vspace{-6mm}  
	\label{fig1}
\end{figure}

Among existing all-in-one restoration paradigms, prompt-based methods have emerged as a representative direction, since they inject degradation-aware control signals into a unified backbone without requiring separate models for different restoration tasks~\cite{34,35,36,38,39,40,41,43}. By introducing degradation-related information through prompt learning~\cite{74}, task adaptation~\cite{59}, contrastive learning~\cite{33}, or other conditional modulation mechanisms~\cite{75}, these methods enable a single network to jointly perform denoising, deraining, dehazing, deblurring, low-light enhancement, and related restoration tasks. As illustrated in \figurename~\ref{fig1}(a), existing prompt-based methods typically generate a single global prompt for the whole image and perform restoration under image-level shared guidance. 

\textbf{Challenges.} In image restoration, degradation types and spatial distributions may vary across regions, and multiple degradations may also coexist in a single image. Although prompt-based all-in-one restoration has achieved encouraging progress, it still faces two challenges in these complex degradation scenarios. \emph{Challenge 1: Fine-grained degradation guidance under spatially varying and mixed degradations.} When different spatial regions exhibit different degradation patterns and restoration needs, relying only on a global prompt makes restoration insufficiently targeted and informative at the query level. Moreover, it remains challenging to construct spatially adaptive queries that accurately capture local degradation characteristics and maintain semantically stable query-wise prototype matching under spatially heterogeneous and mixed degradations. \emph{Challenge 2: Reliable local-global feature aggregation under query-specific guidance.} Even when each spatial query is equipped with degradation-aware guidance, accurate restoration still depends on whether it can effectively aggregate complementary information from local neighborhoods and global contexts. However, because such aggregation is still estimated from degraded features, it may become unreliable for different queries under spatially heterogeneous and mixed degradations, making it difficult to adaptively coordinate local detail modeling and global context modeling.

\textbf{Current attempts.} Existing all-in-one restoration methods have made progress from different perspectives, while most of them focus on different aspects of the problem. On the one hand, methods based on degradation-aware guidance and feature modulation~\cite{33,36,32,42,59} facilitate fine-grained degradation modeling to some extent, but they still mainly rely on image-level shared guidance and thus do not adequately provide query-specific spatially adaptive degradation guidance under spatially heterogeneous and mixed degradations. On the other hand, architectural or routing adaptation~\cite{57,58,44}, as well as feature modulation mechanisms~\cite{36,42,59}, also improve feature interaction and response estimation, but they still largely depend on implicit feature interaction and therefore fall short of reliably modeling local-global complementary aggregation under query-specific guidance. Therefore, a unified mechanism that can jointly address the above limitations remains lacking.

\textbf{Our contributions.} To address the above challenges, we propose QuReC, a unified all-in-one image restoration framework with query-specific guidance and local-global response calibration. As illustrated in \figurename~\ref{fig1}(b), instead of relying on a single image-level prompt, QuReC reconstructs degradation-aware query representations for different spatial locations to provide fine-grained spatially adaptive guidance. On top of this query-specific guidance, QuReC further improves the reliability of local-global feature aggregation. Specifically, to address  \emph{Challenge 1}, we design a Degradation-Guided Query Reconstruction Module (DQRM), which enables each spatial query to adaptively match against a degradation prototype space and reconstruct a corresponding degradation-aware query representation, thereby providing fine-grained spatially adaptive restoration guidance. To further stabilize this query-wise prototype matching process, we introduce a weakly supervised prototype matching learning strategy to improve optimization stability and degradation semantic consistency. To address  \emph{Challenge 2}, we further propose a Local-Global Response Calibration Module (LGRCM), which adopts a local-global dual-branch design to jointly model neighborhood details and global contextual information, while using learnable priors to calibrate the aggregation responses. In this way, DQRM determines what each spatial query should focus on, while LGRCM further improves how each query aggregates restoration information from local and global contexts. Together, they enable more reliable unified restoration under diverse and spatially complex degradation conditions. 

The main contributions are summarized as follows:
\begin{itemize}[leftmargin=*, topsep=1pt, itemsep=1pt, parsep=2pt, partopsep=2pt]
	\item We design a Degradation-Guided Query Reconstruction Module (DQRM) for fine-grained query-specific degradation modeling in all-in-one restoration, together with a weakly supervised prototype matching learning strategy to improve the stability and semantic consistency of query-wise degradation matching.
	\item We design a Local-Global Response Calibration Module (LGRCM) for reliable local-global feature aggregation in unified restoration, improving the coordination between local detail preservation and global structural modeling.
	\item Extensive experiments on multiple all-in-one image restoration benchmarks demonstrate that QuReC achieves superior performance against state-of-the-art methods in both quantitative and qualitative evaluations.
\end{itemize}

\begin{figure*}[t!]
	\centering
	\includegraphics[width=0.82\linewidth]{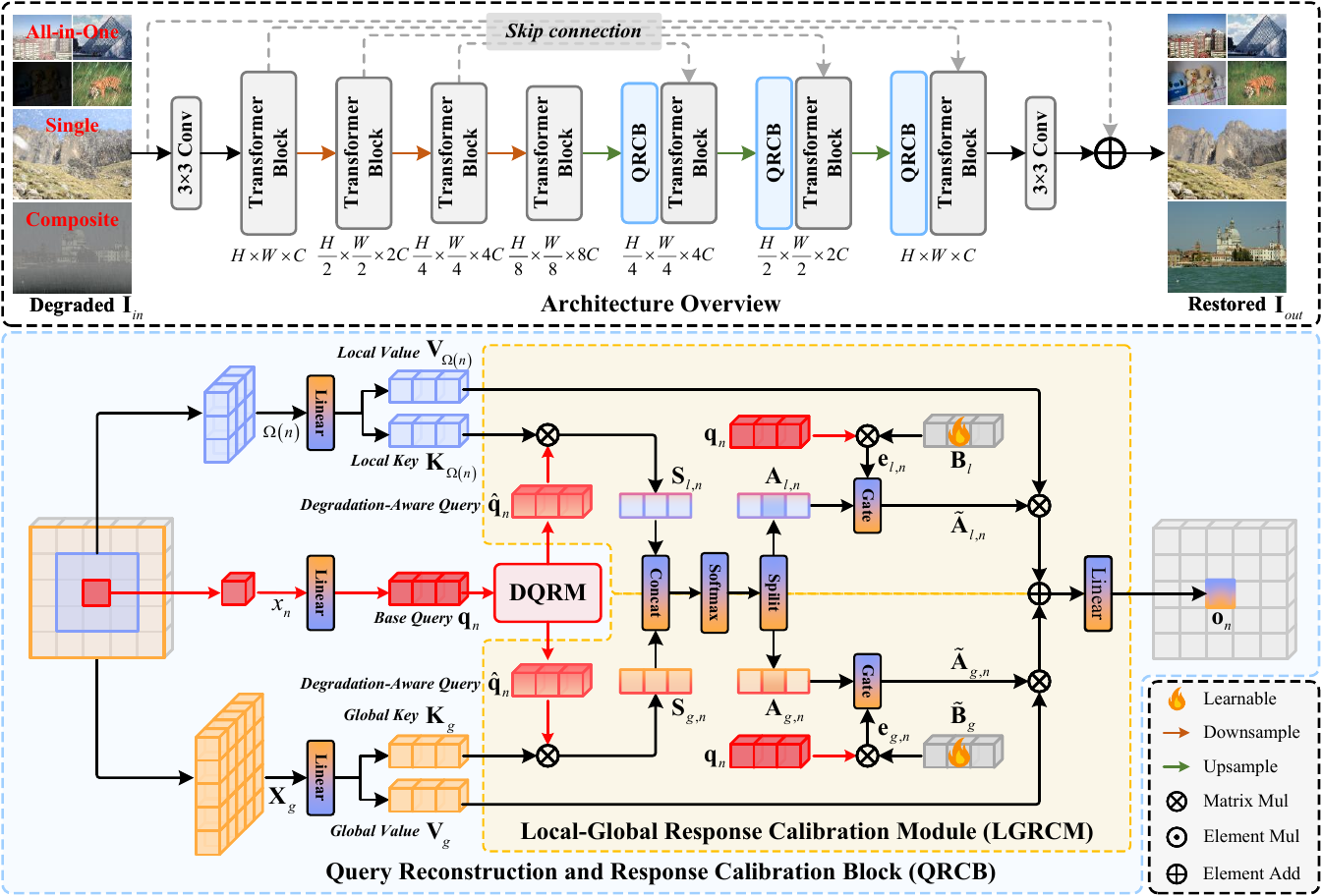}
	\vspace{-2mm}
	\caption{Overview of the proposed QuReC framework. The upper part presents the overall encoder-decoder architecture, in which a Query Reconstruction and Response Calibration Block (QRCB) is inserted into each of the three decoder stages. The lower part illustrates the structure of a QRCB, which consists of a Degradation-Guided Query Reconstruction Module (DQRM) and a Local-Global Response Calibration Module (LGRCM).}
	\vspace{-3mm}
	\label{fig2}
\end{figure*}

\section{Related Work}
\textbf{Single-Task Image Restoration.}
Recovering high-quality images from degraded inputs remains a fundamental problem in computer vision. Early approaches heavily relied on handcrafted priors and empirical features to model complex and uncertain degradation processes. Although these traditional methods can produce favorable results in specific scenarios, they often struggle to generalize to diverse degradations. The advent of deep learning has significantly advanced this field, bringing convolutional neural networks to the forefront \cite{46,47,13,21,49,50,51,52}. A large body of CNN-based methods has been developed for specific restoration tasks, such as denoising \cite{5,53}, deraining \cite{15,16}, and dehazing \cite{21,55}. However, conventional CNNs are inherently limited in modeling long-range dependencies due to their local receptive fields. To address this issue, vision Transformers \cite{54} have been increasingly introduced into image restoration. Representative methods such as Uformer \cite{30}, SwinIR \cite{56}, and Restormer \cite{31} leverage self-attention mechanisms to better capture long-range contextual information and restore complex textures and structures. Despite their strong performance, most existing restoration methods remain task-specific and are typically optimized for a single degradation type. Therefore, although single-task methods have achieved impressive results on individual benchmarks, they generally lack the flexibility to handle diverse degradations within a unified framework.


\textbf{All-in-One Image Restoration.}
To overcome the limitations of single-task image restoration, all-in-one image restoration has emerged as an important research direction, aiming to handle multiple degradation types within a unified network architecture~\cite{37,40,45,57,58, 76,77,78,79,80}. Existing methods generally address this problem from two perspectives, namely arichitectural or routing adaptation and degradation-aware feature modulation. Early studies mainly focused on architectural design and representation learning. For example, BWRAS \cite{57} proposed a single-encoder multi-decoder framework for complex weather degradations, while TKMCR \cite{58} introduced a two-stage knowledge transfer strategy based on a multi-teacher single-student framework. IDR \cite{40} further developed degradation-guided representation learning to facilitate unified restoration. AirNet \cite{33} advanced this direction by learning discriminative degradation representations via contrastive learning for blind restoration. More recently, MoCE-IR \cite{44} introduced a complexity-aware Mixture-of-Experts mechanism to dynamically route different restoration cases through expert specialization. Another prevalent line of research focuses on degradation-aware feature modulation in a shared restoration network. PromptIR \cite{36} implicitly encodes degradation-related information as visual prompts to guide feature reconstruction, while InstructIR \cite{32} further exploits human-written text instructions for multi-task restoration. DFPIR \cite{42} introduces CLIP-encoded text prompts to achieve degradation-aware feature perturbation, and AdaIR \cite{59} explores adaptive frequency modulation to decouple degradation-specific responses from a frequency perspective. These methods have demonstrated that explicit degradation guidance is effective for improving unified restoration performance. Different from existing all-in-one restoration methods, our method focuses on query-wise degradation-guided restoration and response calibration for finer-grained restoration under complex degradations.

\section{Method}
\subsection{Overall Pipeline}
The overall architecture of the proposed framework is illustrated in \figurename~\ref{fig2}. Given a degraded input image $\mathbf{I}_{in}$, our goal is to recover the corresponding clean image $\mathbf{I}_{out}$ using the unified restoration framework QuReC. The model adopts an asymmetric four-stage encoder-decoder architecture. A $3\times3$ convolution first extracts shallow features from $\mathbf{I}_{in}$. The encoder is composed of hierarchically stacked Transformer blocks, which progressively transform high-resolution features into compact latent representations. On the decoder side, in addition to standard Transformer-based reconstruction blocks, we insert three Query Reconstruction and Response Calibration Blocks (QRCBs) at progressively increasing spatial resolutions. Each QRCB consists of two components: a Degradation-Guided Query Reconstruction Module (DQRM) and a Local-Global Response Calibration Module (LGRCM). DQRM reconstructs a degradation-aware query for each spatial location through prototype-based matching, thereby determining what degradation semantics each query should focus on. Based on the reconstructed queries, LGRCM further performs local-global dual-branch aggregation together with prior-guided response calibration, thereby determining how each query reliably aggregates complementary information from local neighborhoods and global contexts. In this way, each QRCB follows a progressive restoration process, namely query-wise degradation specification followed by calibrated response aggregation.

\begin{figure}[t!]
	\centering
	\includegraphics[width=1\linewidth]{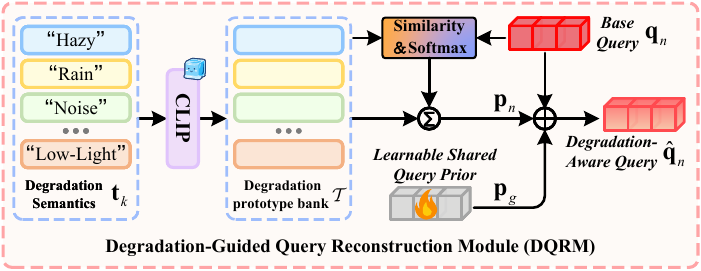}
	\vspace{-6mm}
	\caption{Illustration of the Degradation-Guided Query Reconstruction Module (DQRM).}
	\vspace{-6mm}
	\label{fig3}
\end{figure}

\subsection{Degradation-Guided Query Reconstruction Module}
DQRM is designed to provide query-wise and spatially adaptive degradation guidance. Instead of assigning a single shared prompt to the whole image, DQRM allows each spatial query to adaptively match the most relevant degradation semantics and reconstruct a corresponding degradation-aware query. Specifically, we construct a degradation prototype bank $\mathcal{T}=\{\mathbf{t}_{k}\}_{k=1}^{K}$, where $\mathbf{t}_{k}\in\mathbb{R}^{D}$ denotes the $k$-th degradation prototype embedding. The prototype bank is first built from a set of degradation descriptions according to the training setting and then encoded by a frozen CLIP text encoder to obtain the corresponding prototype embeddings.

Given the input feature sequence $\mathbf{X}\in\mathbb{R}^{N\times C}$, we first project each spatial feature $\mathbf{x}_{n}$ into a prototype-matching space, denoted as $\mathbf{r}_{n}=\phi_{r}(\mathbf{x}_{n})\in\mathbb{R}^{D}$, where $\phi_r(\cdot)$ denotes a learnable projection function. To perform matching, both the projected feature and the prototype embeddings are normalized. The token-specific degradation prompt is then obtained by softly aggregating the prototype bank:
\begin{equation}
	\mathbf{u}_{n}
	=
	\sum_{k=1}^{K}
	\frac{\exp\!\left(\alpha\, \bar{\mathbf{r}}_{n}^{\top}\bar{\mathbf{t}}_{k}\right)}
	{\sum_{j=1}^{K}\exp\!\left(\alpha\, \bar{\mathbf{r}}_{n}^{\top}\bar{\mathbf{t}}_{j}\right)}
	\mathbf{t}_{k},
\end{equation}
where $\bar{\mathbf{r}}_{n}$ and $\bar{\mathbf{t}}_{k}$ denote the normalized projected feature and normalized prototype embedding, respectively, and $\alpha$ is a learnable logit scaling factor.

The resulting prompt is then projected back to the query space as $\mathbf{p}_{n}=\phi_{p}(\mathbf{u}_{n})$, where $\phi_p(\cdot)$ denotes a learnable prompt projection. Meanwhile, we introduce a learnable shared query prior $\mathbf{p}_{g}$, which serves as a task-general anchor shared across different spatial locations. The query-wise degradation prompt $\mathbf{p}_{n}$ provides token-specific degradation semantics, while $\mathbf{p}_{g}$ stabilizes query formation and reduces sensitivity to noisy prototype assignments. Let $\mathbf{q}_{n}=\phi_q(\mathbf{x}_{n})$ denote the base query of the $n$-th token. The final degradation-aware query is reconstructed by combining the base query, the shared query prior, and the token-specific degradation prompt:
\begin{equation}
	\hat{\mathbf{q}}_{n} = \mathbf{q}_{n} + \mathbf{p}_{g} + \mathbf{p}_{n}.
\end{equation}

Unlike image-level prompt assignment, DQRM allows different spatial queries to match different degradation semantics, thereby enabling fine-grained modeling of spatially heterogeneous and compound degradations. For clarity, the above formulation is presented for the $n$-th query, while the same matching operation is applied to all spatial queries in parallel. \figurename~\ref{fig3} provides a schematic illustration of the proposed DQRM.

\textbf{Weakly supervised prototype matching learning.}
Since query-wise prototype matching lacks explicit token-level supervision, optimizing it solely with restoration supervision may lead to prototype collapse or semantically unstable assignments. To stabilize this matching process, we introduce a weakly supervised prototype matching learning strategy consisting of a mixed load balancing loss and a weak matching supervision loss.

Let $N_b$ denote the total number of spatial queries involved in prototype matching for image $b$, aggregated over all QRCB modules, and let $\pi_{b,n,k}$ denote the matching probability of the $n$-th query in image $b$ over the $k$-th degradation prototype. We first compute the image-level average matching distribution and the corresponding soft prototype usage as:
\begin{equation}
	\bar{\pi}_{b,k}=\frac{1}{N_b}\sum_{n=1}^{N_b}\pi_{b,n,k},
	\qquad
	u_k^{\mathrm{soft}}=\frac{1}{B}\sum_{b=1}^{B}\bar{\pi}_{b,k},
\end{equation}
where $\bar{\pi}_{b,k}$ denotes the average matching probability of prototype $k$ in image $b$, and $u_k^{\mathrm{soft}}$ reflects the overall prototype utilization pattern at the probability level. For the hard term, each query is assigned to the prototype with the largest matching probability. We then compute the token-count-weighted hard prototype usage as:
\begin{equation}
	u_k^{\mathrm{hard}}
	=
	\frac{\sum_{b=1}^{B}\sum_{n=1}^{N_b}\mathbb{I}\!\left[\arg\max_j \pi_{b,n,j}=k\right]}
	{\sum_{b=1}^{B} N_b},
\end{equation}
where $\mathbb{I}[\cdot]$ is the indicator function. Unlike the soft term, the hard term reflects the actual token-level discrete occupancy of each prototype after hard assignment. Based on these soft and hard usage statistics, the mixed load balancing loss is defined as:
\begin{equation}
	\mathcal{L}_{\mathrm{bal}} = \alpha_s\!\left(K\sum_{k=1}^{K}(u_k^{\mathrm{soft}})^2-1\right) + \alpha_h\!\left(K\sum_{k=1}^{K}(u_k^{\mathrm{hard}})^2-1\right),
\end{equation}
where $\alpha_s$ and $\alpha_h$ are balancing coefficients. The soft term regularizes prototype usage at the probability level, while the hard term further constrains the discrete assignment pattern. In implementation, the hard assignment term is optimized with a straight-through estimator so that it can regularize prototype matching while preserving gradient flow.

Balanced prototype usage alone does not guarantee semantic correctness. A matcher may still distribute queries over multiple prototypes while drifting away from the degradation semantics present in the input image. We therefore further impose a weak matching supervision loss on the image-level average matching distribution. Instead of using a single one-hot target, we construct an image-level degradation target distribution $\mathbf{y}_b\in\mathbb{R}^{K}$ according to the set of degradation types present in image $b$. Let $\mathcal{S}_b$ denote the set of degradation categories contained in image $b$. The target distribution is defined as:
\begin{equation}
	y_{b,k}=
	\left\{
	\begin{array}{c l}
		\frac{1}{|\mathcal{S}_b|} & ,\ \text{if } k\in\mathcal{S}_b,\\
		0 & ,\ \text{otherwise}.
	\end{array}
	\right.
\end{equation}
Based on this target distribution, the weak matching supervision loss is formulated as:
\begin{equation}
	\mathcal{L}_{\mathrm{match}} =-\frac{1}{B}\sum_{b=1}^{B}\sum_{k=1}^{K} y_{b,k}\log \bar{\pi}_{b,k}.
\end{equation}
This loss does not directly supervise token-level assignments. Instead, it provides a weak semantic constraint on the overall matching distribution, encouraging the activated prototypes to remain consistent with the degradation types present in the image while preserving query-wise flexibility. For clarity, we denote the overall prototype matching learning objective as:
\begin{equation}
	\mathcal{L}_{\mathrm{pm}}
	=
	\lambda_{\mathrm{bal}}\mathcal{L}_{\mathrm{bal}}
	+
	\lambda_{\mathrm{match}}\mathcal{L}_{\mathrm{match}},
\end{equation}
where $\lambda_{\mathrm{bal}}$ and $\lambda_{\mathrm{match}}$ are the corresponding weighting coefficients. The two regularization terms play complementary roles. The mixed load balancing loss encourages balanced prototype utilization from both probabilistic and discrete perspectives, while the weak matching supervision loss aligns the overall matching distribution with the image-level degradation semantics. Together, they make query-wise prototype matching more stable, more interpretable, and more effective for all-in-one image restoration.

%
\subsection{Local-Global Response Calibration Module}
Although DQRM provides query-specific degradation-aware guidance for different spatial locations, accurate restoration still requires each query to reliably aggregate complementary information from local neighborhoods and global contexts. To address this issue, we propose the Local-Global Response Calibration Module (LGRCM), which performs local-global dual-branch aggregation together with prior-guided calibration.

Given the input feature sequence $\mathbf{X}\in\mathbb{R}^{N\times C}$, we construct two aggregation branches for each reconstructed query: a local branch and a global branch. For the $n$-th query, let $\Omega(n)$ denote the set of tokens within the $k\times k$ local window centered at that query location, and let $\mathbf{X}_g\in\mathbb{R}^{N_g\times C}$ denote a spatially reduced global feature sequence derived from $\mathbf{X}$, where $N_g$ is the reduced sequence length. The local keys and values are obtained by applying linear projections to the features within the local window $\Omega(n)$, whereas the global keys and values are obtained by applying linear projections to the reduced global sequence $\mathbf{X}_g$. The branch-wise attention logits are then computed using the reconstructed query $\hat{\mathbf{q}}_n$:
\begin{equation}
	\mathbf{S}_{l,n} = \hat{\mathbf{q}}_n \mathbf{K}_{\Omega(n)}^\top,\qquad
	\mathbf{S}_{g,n} = \hat{\mathbf{q}}_n \mathbf{K}_{g}^\top,
\end{equation}
where $\mathbf{K}_{\Omega(n)}$ and $\mathbf{K}_{g}$ denote the local and global keys, respectively. The resulting local and global logits are not normalized independently; instead, they are concatenated and jointly normalized by a shared Softmax:
\begin{equation}
	\mathbf{A}_{l,n},\mathbf{A}_{g,n}
	=
	\mathrm{Split}\!\left(
	\mathrm{Softmax}\!\left(
	\mathrm{Concat}\!\left(
	\mathbf{S}_{l,n},\mathbf{S}_{g,n}
	\right)
	\right)
	\right).
\end{equation}
This joint normalization places the local and global branches in a shared attention space, enabling the model to adaptively coordinate local detail modeling and global context modeling under different degradation conditions.

Although the dynamic attention weights $\mathbf{A}_{l,n}$ and $\mathbf{A}_{g,n}$ determine where features should be aggregated, they are still estimated from degraded content and may become unreliable. To improve robustness, we introduce two learnable prior banks for calibration, namely a local prior bank $\mathbf{B}_{l}$ and a global prior bank $\mathbf{B}_{g}$. Since the global branch operates on a reduced token sequence, the global prior bank is resized to the current pooled resolution before retrieval.

Importantly, prior retrieval is performed using the base query $\mathbf{q}_{n}$ rather than the reconstructed query $\hat{\mathbf{q}}_{n}$. This design decouples degradation-aware query reconstruction from aggregation calibration: $\hat{\mathbf{q}}_{n}$ is responsible for degradation-aware correspondence modeling in attention computation, whereas $\mathbf{q}_{n}$ is used to retrieve relatively task-agnostic structural priors shared across restoration tasks. The resulting prior responses for the $n$-th query are given by:
\begin{equation}
	\mathbf{e}_{l,n}=\mathbf{q}_{n}\mathbf{B}_{l},\qquad
	\mathbf{e}_{g,n}=\mathbf{q}_{n}\tilde{\mathbf{B}}_{g},
\end{equation}
where $\tilde{\mathbf{B}}_{g}$ denotes the resized global prior bank. These prior responses are then converted into branch-wise gates to modulate the attention-guided aggregation process:
\begin{equation}
	\begin{aligned}
		\mathbf{g}_{l,n}=1+\tanh(\gamma_{l})\odot\tanh(\mathbf{e}_{l,n}),\\
		\mathbf{g}_{g,n}=1+\tanh(\gamma_{g})\odot\tanh(\mathbf{e}_{g,n}).
	\end{aligned}
\end{equation}
where $\gamma_{l}$ and $\gamma_{g}$ are learnable scaling parameters initialized to zero. Owing to the bounded $\tanh(\cdot)$ formulation, the resulting gates provide stable calibration. The branch-wise attention weights are then modulated as $\tilde{\mathbf{A}}_{l,n}=\mathbf{A}_{l,n}\odot\mathbf{g}_{l,n}$ and $\tilde{\mathbf{A}}_{g,n}=\mathbf{A}_{g,n}\odot\mathbf{g}_{g,n}$. Based on these calibrated attention weights, the output of LGRCM for the $n$-th query is computed as:
\begin{equation}
	\mathbf{o}_{n}=\phi_o\!\left(\tilde{\mathbf{A}}_{l,n}\mathbf{V}_{\Omega(n)}+\tilde{\mathbf{A}}_{g,n}\mathbf{V}_{g}\right),
\end{equation}
where $\mathbf{V}_{\Omega(n)}$ denotes the local values associated with the neighborhood $\Omega(n)$, $\mathbf{V}_{g}$ denotes the global values computed from the reduced global sequence $\mathbf{X}_g$, and $\phi_o(\cdot)$ denotes the output mapping function.


\begin{table*}[t]
	\centering
	\small
	\setlength{\tabcolsep}{7pt}
	\renewcommand{\arraystretch}{0.87}
	\caption{Performance Comparison of Different Methods on Three Degradation Tasks. The best and second-best performances for each metric are highlighted with \colorbox[HTML]{ffc7ce}{Red} and \colorbox[HTML]{DFE8FE}{Blue} backgrounds.}
	\vspace{-3mm}  
	\label{tab1}  
	\begin{tabular}{c|*{5}{c c|}c c|c} 
		\toprule
		
		\multicolumn{1}{c|}{\multirow{3}{*}[-1.5ex]{\textbf{Method}}} & \multicolumn{6}{c|}{\textbf{Denoising}} & \multicolumn{2}{c|}{\textbf{Deraining}} & \multicolumn{2}{c|}{\textbf{Dehazing}} & \multicolumn{2}{c|}{\multirow{2}{*}[-0.5ex]{\textbf{Average}}} & \multicolumn{1}{c}{\multirow{3}{*}[-1.5ex]{\textbf{Params (M)}}} \\
		
		\cmidrule{2-11}
		
		& \multicolumn{2}{c|}{$\mathrm{BSD68}_{\sigma=15}$} & \multicolumn{2}{c|}{$\mathrm{BSD68}_{\sigma=25}$} & \multicolumn{2}{c|}{$\mathrm{BSD68}_{\sigma=50}$} & \multicolumn{2}{c|}{Rain100L} & \multicolumn{2}{c|}{SOTS} & \multicolumn{2}{c|}{} & \\
		
		\cmidrule{2-13}
		
		& PSNR & SSIM & PSNR & SSIM & PSNR & SSIM & PSNR & SSIM & PSNR & SSIM & PSNR & SSIM & \\
		\midrule
		
		MPRNet~\cite{68} & 33.27 & 0.920 & 30.76 & 0.871 & 27.29 & 0.761 & 33.86 & 0.958 & 28.00 & 0.958 & 30.63 & 0.894 & 20.10 \\
		
		Restormer~\cite{31} & 33.72 & 0.865 & 30.67 & 0.865 & 27.63 & 0.792 & 33.78 & 0.958 & 27.78 & 0.958 & 30.75 & 0.901 & 26.13 \\
		
		AirNet~\cite{33} & 33.92 & 0.932 & 31.26 & 0.888 & 28.00 & 0.797 & 34.90 & 0.967 & 27.94 & 0.962 & 31.20 & 0.910 & 8.93 \\
		
		
		PromptIR~\cite{36} & 33.98 & 0.933 & 31.31 & 0.888 & 28.06 & 0.799 & 36.37 & 0.972 & 30.58 & 0.974 & 32.06 & 0.913 & 32.96 \\
		
		InstructIR~\cite{32} & \cellcolor[HTML]{DFE8FE}{34.15} & 0.933 & 31.52 & 0.890 & 28.30 & \cellcolor[HTML]{DFE8FE}{0.804} & 37.98 & 0.978 & 30.22 & 0.959 & 32.43 & 0.913 & 15.84 \\
		
		Perceive-IR~\cite{41} & 34.13 & \cellcolor[HTML]{DFE8FE}{0.934} & \cellcolor[HTML]{DFE8FE}{31.53} & 0.890 & \cellcolor[HTML]{DFE8FE}{28.31} & \cellcolor[HTML]{DFE8FE}{0.804} & 38.29 & 0.980 & 30.87 & 0.975 & 32.63 & 0.917 & 42.02 \\
		
		NDR-Restore~\cite{re_1} & 34.01 & 0.932 & 31.36 & 0.887 & 28.10 & 0.798 & 35.42 & 0.969 & 28.64 & 0.962 & 31.51 & 0.910 & 28.40 \\
		
		AdaIR~\cite{59} & 34.12 & \cellcolor[HTML]{ffc7ce}{0.935} & 31.45 & \cellcolor[HTML]{DFE8FE}{0.892} & 28.19 & 0.802 & 38.64 & 0.983 & 31.06 & \cellcolor[HTML]{DFE8FE}{0.980} & 32.69 & 0.918 & 28.77 \\
		
		MoCE-IR~\cite{44} & 34.11 & 0.932 & 31.45 & 0.888 & 28.18 & 0.800 & 38.57 & \cellcolor[HTML]{DFE8FE}{0.984} & 31.34 & 0.979 & 32.73 & 0.917 & 25.35 \\
		
		DFPIR~\cite{42} & 34.14 & \cellcolor[HTML]{ffc7ce}{0.935} & 31.47 & \cellcolor[HTML]{ffc7ce}{0.893} & 28.25 & \cellcolor[HTML]{ffc7ce}{0.806} & \cellcolor[HTML]{DFE8FE}{38.65} & 0.982 & \cellcolor[HTML]{DFE8FE}{31.87} & \cellcolor[HTML]{DFE8FE}{0.980} & \cellcolor[HTML]{DFE8FE}{32.88} & \cellcolor[HTML]{DFE8FE}{0.919} & 31.10 \\
		
		\midrule
		
		QuReC (Ours) & \cellcolor[HTML]{ffc7ce}{34.26} & \cellcolor[HTML]{DFE8FE}{0.934} & \cellcolor[HTML]{ffc7ce}{31.60} & \cellcolor[HTML]{DFE8FE}{0.892} & \cellcolor[HTML]{ffc7ce}{28.34} & \cellcolor[HTML]{ffc7ce}{0.806} & \cellcolor[HTML]{ffc7ce}{38.80} & \cellcolor[HTML]{ffc7ce}{0.985} & \cellcolor[HTML]{ffc7ce}{32.77} & \cellcolor[HTML]{ffc7ce}{0.985} & \cellcolor[HTML]{ffc7ce}{33.16} & \cellcolor[HTML]{ffc7ce}{0.920} & 29.61 \\
		
		\bottomrule
	\end{tabular}
	\vspace{-3mm}  
\end{table*}

\begin{figure*}[t]
	\centering
	\includegraphics[width=0.94\linewidth]{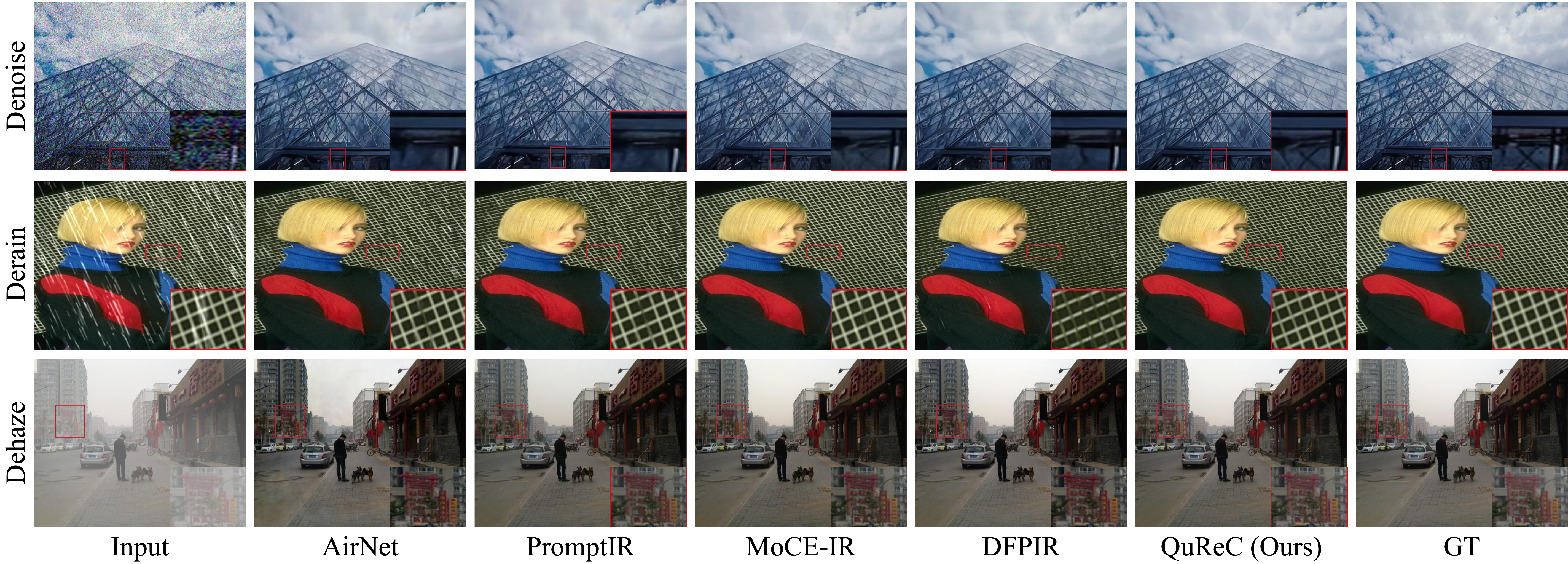}  
	\vspace{-3mm}  
	\caption{Qualitative comparisons of different methods under the all-in-one setting with three degradations. QuReC produces cleaner restoration results while better preserving structural details and visual fidelity.}
	\vspace{-4mm}  
	\label{fig4}
\end{figure*}

\section{Experiments}
\subsection{Experimental Settings}
\textbf{Datasets.} Following the settings of previous studies, we adopted corresponding datasets for different image restoration tasks to evaluate the performance of the proposed model. For image dehazing, we selected the SOTS subset from the RESIDE~\cite{60} dataset. For image deraining, we used the Rain100L~\cite{61} dataset. For image denoising, noisy images were synthetically generated by adding Gaussian noise with noise levels of $\sigma \in \{15, 25, 50\}$ to clean images from the BSD400~\cite{62} and WED~\cite{63} datasets, and evaluation was conducted on the BSD68~\cite{64} dataset. For image deblurring, we employed the GoPro~\cite{65} dataset. For low-light enhancement, we used the LOL~\cite{66} dataset. For the composite degradation setting, experiments were conducted on the CDD11~\cite{67} dataset. This dataset contains 11 degradation scenarios generated from four basic degradation types: Low (L), Haze (H), Rain (R), and Snow (S), either individually or in combination, covering a wide range of typical cases from single degradation to multiple composite degradations. Specifically, the 11 degradation types include L, H, R, S, L + H, L + R, L + S, H + R, H + S, L + H + R, and L + H + S. We conducted extensive experiments on these datasets under three image restoration scenarios: (a) single-degradation, (b) all-in-one, and (c) composite degradation. \textbf{More experimental results are provided in the supplementary material.}

\textbf{Implementation Details.} All experiments were conducted on two NVIDIA GeForce RTX 3090 GPUs. We used the Adam optimizer~\cite{73} with an initial learning rate of $2 \times 10^{-4}$, which was updated using a cosine annealing schedule. During training, images were randomly cropped to $128 \times 128$, and horizontal and vertical flipping were adopted for data augmentation. For the single-degradation, all-in-one, and composite-degradation settings, all models were trained for 150 epochs. In the all-in-one setting, the datasets of different degradation types were merged for joint training under the three-degradation and five-degradation settings. For the composite-degradation setting, experiments were conducted on the CDD11~\cite{67} dataset. 
\begin{table*}[t!]
	\centering
	\small
	\setlength{\tabcolsep}{7pt}
	\renewcommand{\arraystretch}{0.87}
	\caption{Performance Comparison of Different Methods on Five Degradation Tasks. The best and second-best performances for each metric are highlighted with \colorbox[HTML]{ffc7ce}{Red} and \colorbox[HTML]{DFE8FE}{Blue} backgrounds.}
	\vspace{-3mm}  
	\label{tab2} 
	\begin{tabular}{c|*{6}{c c|}c} 
		\toprule
		
		\multicolumn{1}{c|}{\multirow{3}{*}[-1.5ex]{\textbf{Method}}} & \multicolumn{2}{c|}{\textbf{Denoising}} & \multicolumn{2}{c|}{\textbf{Deraining}} & \multicolumn{2}{c|}{\textbf{Dehazing}} & \multicolumn{2}{c|}{\textbf{Deblurring}} & \multicolumn{2}{c|}{\textbf{Low-light}} & \multicolumn{2}{c|}{\multirow{2}{*}[-0.5ex]{\textbf{Average}}} & \multicolumn{1}{c}{\multirow{3}{*}[-1.5ex]{\textbf{Params (M)}}} \\
		
		\cmidrule{2-11}
		
		& \multicolumn{2}{c|}{$\mathrm{BSD68}_{\sigma=25}$} & \multicolumn{2}{c|}{Rain100L} & \multicolumn{2}{c|}{SOTS} & \multicolumn{2}{c|}{GoPro} & \multicolumn{2}{c|}{LOL} & \multicolumn{2}{c|}{} & \\
		
		\cmidrule{2-13}
		
		& PSNR & SSIM & PSNR & SSIM & PSNR & SSIM & PSNR & SSIM & PSNR & SSIM & PSNR & SSIM & \\
		\midrule
		
		NAFNet~\cite{69} & 31.02 & 0.883 & 35.56 & 0.967 & 25.23 & 0.929 & 26.53 & 0.808 & 20.49 & 0.809 & 27.76 & 0.881 & 17.10 \\
		
		Restormer~\cite{31} & \cellcolor[HTML]{DFE8FE}{31.49} & 0.884 & 34.81 & 0.962 & 24.09 & 0.927 & 27.22 & 0.829 & 20.41 & 0.806 & 27.60 & 0.881 & 26.13 \\
		
		AirNet~\cite{33} & 30.91 & 0.882 & 32.98 & 0.951 & 21.04 & 0.884 & 24.35 & 0.781 & 18.18 & 0.735 & 25.49 & 0.846 & 8.93 \\
		
		
		PromptIR~\cite{36} & 31.47 & 0.886 & 36.37 & 0.970 & 26.54 & 0.949 & 28.71 & 0.881 & 22.68 & 0.832 & 29.15 & 0.904 & 32.96 \\
		
		InstructIR~\cite{32} & 31.40 & 0.887 & 36.84 & 0.973 & 27.10 & 0.956 & 29.40 & 0.886 & 23.00 & 0.836 & 29.55 & 0.907 & 15.84 \\
		
		Perceive-IR~\cite{41} & 31.44 & 0.887 & 37.25 & 0.977 & 28.19 & 0.964 & 29.46 & 0.886 & 22.88 & 0.833 & 29.84 & 0.909 & 42.02 \\
		
		AdaIR~\cite{59} & 31.35 & \cellcolor[HTML]{DFE8FE}{0.889} & 38.02 & 0.981 & 30.53 & 0.978 & 28.12 & 0.858 & 23.00 & 0.845 & 30.20 & 0.910 & 28.77 \\
		
		MoCE-IR~\cite{44} & 31.34 & 0.887 & \cellcolor[HTML]{DFE8FE}{38.04} & \cellcolor[HTML]{DFE8FE}{0.982} & 30.48 & 0.974 & \cellcolor[HTML]{DFE8FE}{30.05} & \cellcolor[HTML]{DFE8FE}{0.899} & 23.00 & \cellcolor[HTML]{DFE8FE}{0.852} & 30.58 & \cellcolor[HTML]{DFE8FE}{0.919} & 25.35 \\
		
		DFPIR~\cite{42} & 31.29 & \cellcolor[HTML]{DFE8FE}{0.889} & 37.62 & 0.978 & \cellcolor[HTML]{DFE8FE}{31.64} & \cellcolor[HTML]{DFE8FE}{0.979} & 28.82 & 0.873 & \cellcolor[HTML]{DFE8FE}{23.82} & 0.843 & \cellcolor[HTML]{DFE8FE}{30.64} & 0.913 & 31.10 \\
		\midrule
		
		QuReC (Ours) & \cellcolor[HTML]{ffc7ce}{31.56} & \cellcolor[HTML]{ffc7ce}{0.891} & \cellcolor[HTML]{ffc7ce}{39.23} & \cellcolor[HTML]{ffc7ce}{0.986} & \cellcolor[HTML]{ffc7ce}{31.72} & \cellcolor[HTML]{ffc7ce}{0.982} & \cellcolor[HTML]{ffc7ce}{32.50} & \cellcolor[HTML]{ffc7ce}{0.936} & \cellcolor[HTML]{ffc7ce}{24.23} & \cellcolor[HTML]{ffc7ce}{0.867} & \cellcolor[HTML]{ffc7ce}{31.85} & \cellcolor[HTML]{ffc7ce}{0.932} & 29.61 \\
		
		\bottomrule
	\end{tabular}
	\vspace{-3mm}  
\end{table*}

\begin{table*}[t!]
	\centering
	\small
	\setlength{\tabcolsep}{1pt}
	\renewcommand{\arraystretch}{1.2}
	\caption{Performance Comparison of Different Methods on the CDD11 dataset. The best and second-best performances for each metric are highlighted with \colorbox[HTML]{ffc7ce}{Red} and \colorbox[HTML]{DFE8FE}{Blue} backgrounds.}
	\vspace{-3mm}  
	\label{tab3}  
	\begin{tabularx}{\linewidth}{c|*{11}{>{\centering\arraybackslash}X >{\centering\arraybackslash}X|}>{\centering\arraybackslash}X >{\centering\arraybackslash}X} 
		\toprule
		
		\multicolumn{1}{c|}{\multirow{2}{*}[-0.8ex]{\textbf{Method}}} & \multicolumn{2}{c|}{\textbf{Low (L)}} & \multicolumn{2}{c|}{\textbf{Haze (H)}} & \multicolumn{2}{c|}{\textbf{Rain (R)}} & \multicolumn{2}{c|}{\textbf{Snow (S)}} & \multicolumn{2}{c|}{\textbf{L+H}} & \multicolumn{2}{c|}{\textbf{L+R}} & \multicolumn{2}{c|}{\textbf{L+S}} & \multicolumn{2}{c|}{\textbf{H+R}} & \multicolumn{2}{c|}{\textbf{H+S}} & \multicolumn{2}{c|}{\textbf{L+H+R}} & \multicolumn{2}{c|}{\textbf{L+H+S}} & \multicolumn{2}{c}{\textbf{Average}} \\
		
		\cmidrule{2-25}
		
		& PSNR & SSIM & PSNR & SSIM & PSNR & SSIM & PSNR & SSIM & PSNR & SSIM & PSNR & SSIM & PSNR & SSIM & PSNR & SSIM & PSNR & SSIM & PSNR & SSIM & PSNR & SSIM & PSNR & SSIM \\
		\midrule
		
		AirNet~\cite{33} & 24.83 & 0.778 & 24.21 & 0.951 & 26.55 & 0.891 & 26.79 & 0.919 & 23.23 & 0.779 & 22.82 & 0.710 & 23.29 & 0.723 & 22.21 & 0.868 & 23.29 & 0.901 & 21.80 & 0.708 & 22.24 & 0.725 & 23.75 & 0.814 \\
		
		PromptIR~\cite{36} & 26.32 & 0.805 & 26.10 & 0.969 & 31.56 & 0.946 & 31.53 & 0.960 & 24.49 & 0.789 & 25.05 & 0.771 & 24.51 & 0.761 & 24.54 & 0.924 & 23.70 & 0.925 & 23.74 & 0.752 & 23.33 & 0.747 & 25.90 & 0.850 \\
		
		WGWSNet~\cite{71} & 24.39 & 0.774 & 27.90 & 0.982 & 33.15 & 0.964 & 34.43 & 0.973 & 24.27 & 0.800 & 25.06 & 0.772 & 24.60 & 0.765 & 27.23 & 0.955 & 27.65 & 0.960 & 23.90 & 0.772 & 23.97 & 0.771 & 26.96 & 0.863 \\
		
		WeatherDiff~\cite{72} & 23.58 & 0.763 & 21.99 & 0.904 & 24.85 & 0.885 & 24.80 & 0.888 & 21.83 & 0.756 & 22.69 & 0.730 & 22.12 & 0.707 & 21.25 & 0.868 & 21.99 & 0.868 & 21.23 & 0.716 & 21.04 & 0.698 & 22.49 & 0.799 \\
		
		OneRestore~\cite{67} & 26.48 & \cellcolor[HTML]{DFE8FE}{0.826} & 32.52 & \cellcolor[HTML]{DFE8FE}{0.990} & 33.40 & 0.964 & 34.31 & 0.973 & 25.79 & \cellcolor[HTML]{DFE8FE}{0.822} & 25.58 & 0.799 & 25.19 & 0.789 & \cellcolor[HTML]{DFE8FE}{29.99} & 0.957 & \cellcolor[HTML]{DFE8FE}{30.21} & 0.964 & 24.78 & 0.788 & 24.90 & \cellcolor[HTML]{DFE8FE}{0.791} & 28.47 & 0.878 \\
		
		MoCE-IR~\cite{44} & \cellcolor[HTML]{DFE8FE}{27.26} & 0.824 & \cellcolor[HTML]{DFE8FE}{32.66} & \cellcolor[HTML]{DFE8FE}{0.990} & \cellcolor[HTML]{DFE8FE}{34.31} & \cellcolor[HTML]{DFE8FE}{0.970} & \cellcolor[HTML]{DFE8FE}{35.91} & \cellcolor[HTML]{DFE8FE}{0.980} & \cellcolor[HTML]{DFE8FE}{26.24} & 0.817 & \cellcolor[HTML]{DFE8FE}{26.25} & \cellcolor[HTML]{DFE8FE}{0.800} & \cellcolor[HTML]{DFE8FE}{26.04} & \cellcolor[HTML]{DFE8FE}{0.793} & 29.93 & \cellcolor[HTML]{DFE8FE}{0.964} & 30.19 & \cellcolor[HTML]{DFE8FE}{0.970} & \cellcolor[HTML]{DFE8FE}{25.41} & \cellcolor[HTML]{DFE8FE}{0.789} & \cellcolor[HTML]{DFE8FE}{25.39} & 0.790 & \cellcolor[HTML]{DFE8FE}{29.05} & \cellcolor[HTML]{DFE8FE}{0.881} \\
		\midrule
		QuReC (Ours) & \cellcolor[HTML]{ffc7ce}{27.58} & \cellcolor[HTML]{ffc7ce}{0.833} & \cellcolor[HTML]{ffc7ce}{36.42} & \cellcolor[HTML]{ffc7ce}{0.995} & \cellcolor[HTML]{ffc7ce}{35.71} & \cellcolor[HTML]{ffc7ce}{0.977} & \cellcolor[HTML]{ffc7ce}{37.86} & \cellcolor[HTML]{ffc7ce}{0.985} & \cellcolor[HTML]{ffc7ce}{27.27} & \cellcolor[HTML]{ffc7ce}{0.832} & \cellcolor[HTML]{ffc7ce}{27.05} & \cellcolor[HTML]{ffc7ce}{0.817} & \cellcolor[HTML]{ffc7ce}{26.90} & \cellcolor[HTML]{ffc7ce}{0.812} & \cellcolor[HTML]{ffc7ce}{32.71} & \cellcolor[HTML]{ffc7ce}{0.974} & \cellcolor[HTML]{ffc7ce}{33.29} & \cellcolor[HTML]{ffc7ce}{0.979} & \cellcolor[HTML]{ffc7ce}{26.50} & \cellcolor[HTML]{ffc7ce}{0.809} & \cellcolor[HTML]{ffc7ce}{26.57} & \cellcolor[HTML]{ffc7ce}{0.811} & \cellcolor[HTML]{ffc7ce}{30.71} & \cellcolor[HTML]{ffc7ce}{0.903}\\
		
		\bottomrule
	\end{tabularx}
	\vspace{-3mm}  
\end{table*}

\subsection{Comparison with State-of-the-Art Methods}
\textbf{Comparison on Three Restoration Tasks.}
We first evaluate QuReC on three restoration tasks, including denoising, deraining, and dehazing, and compare it with both general restoration models (MPRNet~\cite{68} and Restormer~\cite{31}) and recent all-in-one methods (AirNet~\cite{33}, PromptIR~\cite{36}, InstructIR~\cite{32}, Perceive-IR~\cite{41}, NDR-Restore~\cite{re_1}, AdaIR~\cite{59}, MoCE-IR~\cite{44}, and DFPIR~\cite{42}). As reported in \tablename~\ref{tab1}, QuReC achieves the best PSNR in all five evaluation settings and obtains the highest average PSNR and SSIM, demonstrating the strongest overall performance. Compared with DFPIR, the previous best-performing method on average, QuReC improves the average PSNR by 0.28 dB. In particular, QuReC yields PSNR gains of 0.12 dB, 0.13 dB, and 0.09 dB on BSD68 with noise levels 15, 25, and 50, respectively, and further improves the performance by 0.15 dB on Rain100L and 0.90 dB on SOTS. Although it does not achieve the best SSIM in some denoising settings, QuReC still delivers the best overall performance across different tasks, verifying its effectiveness in unified restoration.

\textbf{Comparison on Five Degradation Tasks.}
We further extend the evaluation to five restoration tasks, including denoising, deraining, dehazing, deblurring, and low-light enhancement. The compared methods include two general restoration models (NAFNet~\cite{69} and Restormer~\cite{31}) and several recent all-in-one methods (AirNet~\cite{33}, PromptIR~\cite{36}, InstructIR~\cite{32}, Perceive-IR~\cite{41}, AdaIR~\cite{59}, MoCE-IR~\cite{44}, and DFPIR~\cite{42}). As shown in \tablename~\ref{tab2}, QuReC achieves the best PSNR and SSIM in all five evaluation settings and also attains the highest average PSNR and SSIM. Compared with DFPIR, QuReC improves the average PSNR by 1.21 dB. The gain is especially prominent on GoPro deblurring, where QuReC surpasses DFPIR by 3.68 dB, while also improving the performance by 1.61 dB on Rain100L, 0.41 dB on LOL, 0.27 dB on BSD68 (\(\sigma=25\)), and 0.08 dB on SOTS. These results show that QuReC generalizes well to a broader range of degradations and remains effective as the task space becomes more challenging.

\begin{table*}[t]
	\centering
	\small
	\renewcommand{\arraystretch}{1}
	
	\begin{minipage}[t]{0.33\textwidth}
		\centering
		\setlength{\tabcolsep}{3pt}
		\renewcommand{\arraystretch}{0.97}
		\captionof{table}{Overall ablation study of QuReC.}
		\vspace{-3mm}  
		\label{tab4}
		\begin{tabular}{c|c c c|c c}
			\toprule
			Setting & DQRM & LGRCM & $\mathcal{L}_{\mathrm{pm}}$ & PSNR & SSIM \\
			\midrule
			(a) & \xmark & \xmark & \xmark & 30.75 & 0.901 \\
			(b) & \cmark & \xmark & \xmark & 31.84 & 0.909 \\
			(c) & \xmark & \cmark & \xmark & 31.98 & 0.910 \\
			(d) & \cmark & \cmark & \xmark & \cellcolor[HTML]{DFE8FE}{32.78} & \cellcolor[HTML]{DFE8FE}{0.916} \\
			(e) & \cmark & \cmark & \cmark & \cellcolor[HTML]{ffc7ce}{33.16} & \cellcolor[HTML]{ffc7ce}{0.920} \\
			\bottomrule
		\end{tabular}
	\end{minipage}
	\hfill
	\begin{minipage}[t]{0.32\textwidth}
		\centering
		\setlength{\tabcolsep}{5pt}
		\renewcommand{\arraystretch}{1.17}
		\captionof{table}{Ablation study of DQRM.}
		\vspace{-3mm}  
		\label{tab5}
		\begin{tabular}{c|c c c|c c}
			\toprule
			Setting & QW & $\mathbf{p}_{g}$ & $\mathcal{L}_{\mathrm{pm}}$ & PSNR & SSIM \\
			\midrule
			(a) & \xmark & \xmark & \xmark & 32.26 & 0.913 \\
			(b) & \cmark & \xmark & \xmark & 32.56 & 0.915 \\
			(c) & \cmark & \cmark & \xmark & \cellcolor[HTML]{DFE8FE}{32.78} & \cellcolor[HTML]{DFE8FE}{0.916} \\
			(d) & \cmark & \cmark & \cmark & \cellcolor[HTML]{ffc7ce}{33.16} & \cellcolor[HTML]{ffc7ce}{0.920} \\
			\bottomrule
		\end{tabular}
	\end{minipage}
	\hfill
	\begin{minipage}[t]{0.32\textwidth}
		\centering
		\setlength{\tabcolsep}{5pt}
		\renewcommand{\arraystretch}{0.97}
		\captionof{table}{Ablation study of LGRCM.}
		\vspace{-3mm}  
		\label{tab7}
		\begin{tabular}{c|c c c c|c c}
			\toprule
			Setting & LB & GB & JS & PC & PSNR & SSIM \\
			\midrule
			(a) & \cmark & \xmark & \xmark & \xmark & 32.37 & 0.913 \\
			(b) & \xmark & \cmark & \xmark & \xmark & 32.49 & 0.914 \\
			(c) & \cmark & \cmark & \xmark & \xmark & 32.82 & 0.916 \\
			(d) & \cmark & \cmark & \cmark & \xmark & \cellcolor[HTML]{DFE8FE}{32.97} & \cellcolor[HTML]{DFE8FE}{0.918} \\
			(e) & \cmark & \cmark & \cmark & \cmark & \cellcolor[HTML]{ffc7ce}{33.16} & \cellcolor[HTML]{ffc7ce}{0.920} \\
			\bottomrule
		\end{tabular}
	\end{minipage}
\end{table*}

\textbf{Comparison on Composite Degradations.}
We further evaluate QuReC on the challenging CDD11 benchmark and compare it with several recent all-in-one restoration methods, including AirNet~\cite{33}, PromptIR~\cite{36}, WGWSNet~\cite{71}, WeatherDiff~\cite{72}, OneRestore~\cite{67}, and MoCE-IR. As shown in \tablename~\ref{tab3}, QuReC achieves the best PSNR and SSIM in all 11 evaluation settings, as well as the highest average PSNR and SSIM. Compared with MoCE-IR, which previously achieved the best overall performance, QuReC improves the average PSNR and SSIM by 1.66 dB and 0.022, respectively. Notably, QuReC brings consistent gains on both single and composite degradation settings, including improvements of 0.32 dB, 3.76 dB, 1.40 dB, and 1.95 dB on the four single-degradation settings of low-light, haze, rain, and snow, respectively. It further achieves clear gains of 1.03 dB, 0.80 dB, 0.86 dB, 2.72 dB, 3.08 dB, 1.09 dB, and 1.18 dB on the composite settings of L+H, L+R, L+S, H+R, H+S, L+H+R, and L+H+S, respectively. These results demonstrate the strong robustness of QuReC under compositionally diverse and highly challenging degradation scenarios.

\textbf{Visual Results.}
Figure~\ref{fig4} presents qualitative comparisons on three representative restoration tasks, including denoising, deraining, and dehazing. Compared with previous all-in-one restoration methods, QuReC consistently produces cleaner and more faithful results across all cases. In the denoising example, competing methods either leave noticeable residual noise or oversmooth fine structures, whereas QuReC better preserves sharp edges and structural details in the highlighted region. In the deraining case, existing methods tend to blur the underlying lattice texture or fail to remove rain streaks completely, while QuReC restores clearer structural patterns with fewer artifacts. In the dehazing example, other methods still suffer from limited contrast enhancement and degraded distant details, whereas our method recovers clearer scene structures and more natural visibility. Overall, these qualitative comparisons show that QuReC achieves a better trade-off between degradation removal and detail preservation across different restoration tasks.
\begin{figure}[t!]
	\centering
	\includegraphics[width=1\linewidth]{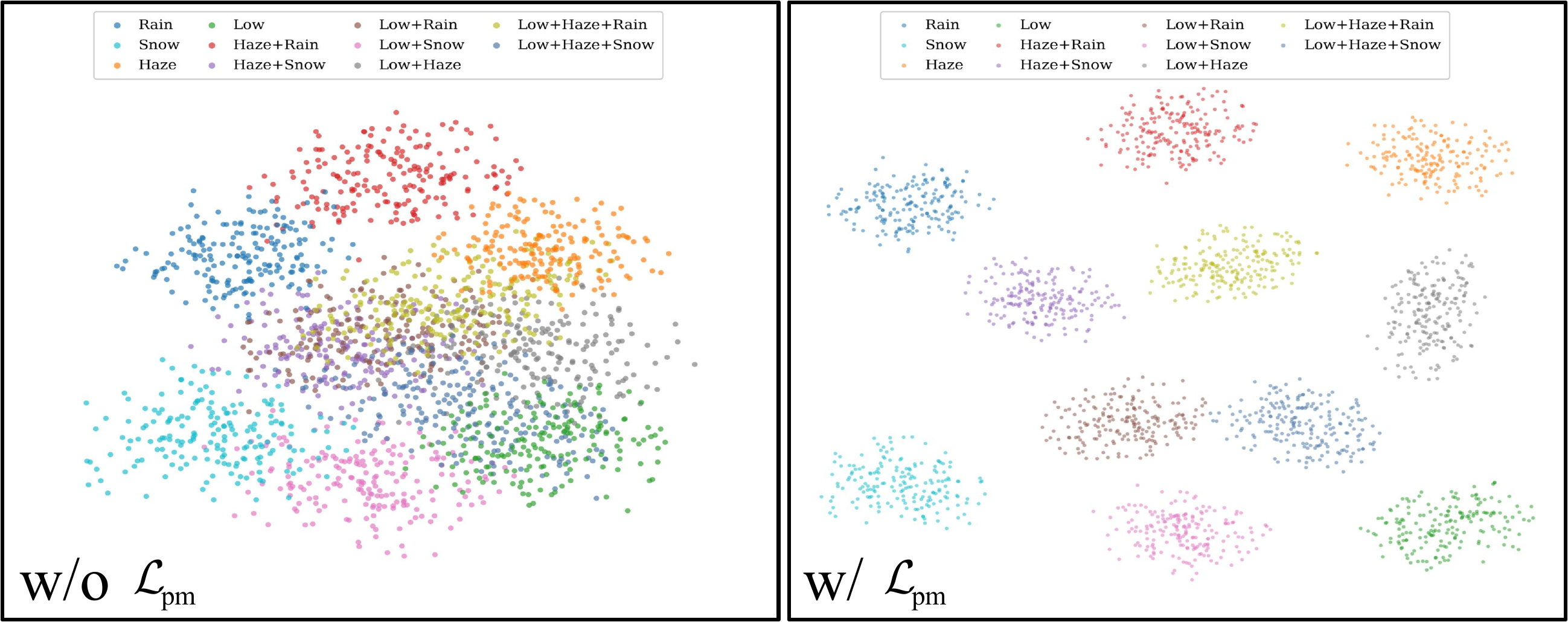}
	\vspace{-6mm}  
	\caption{The t-SNE visualization of routing logits shows that $\mathcal{L}_{\mathrm{pm}}$ makes the prototype matching space more structured and more separable across different degradation types.}
	\vspace{-4mm}  
	\label{fig5}
\end{figure}
\subsection{Ablation Study}
We perform ablation experiments to validate the effectiveness of the proposed framework and its main components. \tablename~\ref{tab4} reports the overall ablation results, while \tablename~\ref{tab5} and \tablename~\ref{tab7} further analyze DQRM and LGRCM, respectively. For brevity, in \tablename~\ref{tab5}, QW and $\mathbf{p}_{g}$ denote query-wise prompting and the shared query prior, respectively. In \tablename~\ref{tab7}, LB, GB, JS, and PC denote the local branch, global branch, joint Softmax, and prior calibration, respectively.

\textbf{Effectiveness of the Proposed Components.}
\tablename~\ref{tab4} shows that each proposed component consistently improves restoration performance. Starting from the baseline, introducing DQRM alone improves the average performance from 30.75/0.901 to 31.84/0.909 in PSNR/SSIM, indicating that query-wise degradation-guided query reconstruction is effective for unified restoration. Replacing DQRM with LGRCM also yields a clear gain, reaching 31.98/0.910, which verifies the importance of improving aggregation reliability through local-global response modeling. When DQRM and LGRCM are jointly enabled, the performance further rises to 32.78/0.916, suggesting that the two modules are highly complementary. Finally, adding the prototype matching learning objective $\mathcal{L}_{\mathrm{pm}}$ pushes the performance to 33.16/0.920, demonstrating that stable prototype matching learning is crucial for fully exploiting the proposed query-wise degradation guidance.

\textbf{Effectiveness of DQRM.}
\tablename~\ref{tab5} further evaluates the contribution of the key elements in DQRM. Replacing image-level shared prompting with the proposed query-wise prompting improves the average performance from 32.26/0.913 to 32.56/0.915, showing that degradation guidance should be assigned at the query level rather than globally shared across the image. Introducing the shared query prior $\mathbf{p}_{g}$ further improves the performance to 32.78/0.916, suggesting that $\mathbf{p}_{g}$ provides a stable task-general anchor for query reconstruction. When $\mathcal{L}_{\mathrm{pm}}$ is additionally applied, the performance further increases to 33.16/0.920. This result verifies that the proposed prototype matching learning objective effectively stabilizes query-wise prototype assignment and improves the semantic consistency of degradation-aware query reconstruction. To further analyze the effect of $\mathcal{L}_{\mathrm{pm}}$, we visualize the routing logits of the third QRCB in the decoder using t-SNE. As shown in \figurename~\ref{fig5}, without $\mathcal{L}_{\mathrm{pm}}$, different degradation types are heavily entangled in the prototype matching space. After introducing $\mathcal{L}_{\mathrm{pm}}$, the routing space becomes much more structured and exhibits clearer degradation-dependent clustering. This observation further verifies that $\mathcal{L}_{\mathrm{pm}}$ effectively regularizes the query-wise prototype matching process and improves degradation semantic consistency.

\textbf{Effectiveness of LGRCM.}
\tablename~\ref{tab7} evaluates the effectiveness of the design choices in LGRCM. Using only the local branch achieves 32.37/0.913, while using only the global branch yields 32.49/0.914, indicating that global contextual modeling is slightly more beneficial in the unified restoration setting. Combining the two branches further improves the performance to 32.82/0.916, confirming their complementary roles in local detail aggregation and global context modeling. On top of the dual-branch design, introducing JS further boosts the performance to 32.97/0.918, demonstrating that placing local and global responses in a shared competitive normalization space is more effective than treating them independently. Finally, enabling PC achieves the best result of 33.16/0.920, which shows that prior-guided calibration further enhances the reliability of attention-guided aggregation.

\section{Conclusion}
In this paper, we proposed QuReC, a unified framework for all-in-one image restoration. Unlike existing prompt-based restoration methods that mainly rely on image-level shared guidance, QuReC performs restoration through query-wise degradation-guided query reconstruction and local-global response calibration. Specifically, DQRM reconstructs query-specific degradation-aware representations through adaptive prototype matching, while LGRCM improves the reliability of feature aggregation by coordinating local-global aggregation with prior-guided calibration. Extensive experiments on three-degradation, five-degradation, and composite-degradation benchmarks demonstrate that QuReC consistently outperforms existing methods. In future work, we will explore more general degradation modeling and more scalable unified restoration frameworks for more challenging degradation scenarios.

\begin{acks}
	This work is supported by Science and Technology Major Special Program of Jiangsu Province under Grant No. BG2024028; Jiangsu Provincial Frontier Technology Research and Development Program under Grant No. BF2024070; the Fundamental Research Funds for the Central Universities; National Natural Science Foundation of China under Grant Nos. 62472094, 62572119, 62232004; Jiangsu Provincial Key Laboratory of Network and Information Security under Grant No. BM2003201; Key Laboratory of Computer Network and Information Integration (Ministry of Education, China) under Grant No. 93K-9; Collaborative Innovation Center of Novel Software Technology and Industrialization; Collaborative Innovation Center of Wireless Communications Technology.
\end{acks}

\bibliographystyle{ACM-Reference-Format}
\bibliography{sample-base}


\begin{thebibliography}{86}


\ifx \showCODEN    \undefined \def \showCODEN     #1{\unskip}     \fi
\ifx \showISBNx    \undefined \def \showISBNx     #1{\unskip}     \fi
\ifx \showISBNxiii \undefined \def \showISBNxiii  #1{\unskip}     \fi
\ifx \showISSN     \undefined \def \showISSN      #1{\unskip}     \fi
\ifx \showLCCN     \undefined \def \showLCCN      #1{\unskip}     \fi
\ifx \shownote     \undefined \def \shownote      #1{#1}          \fi
\ifx \showarticletitle \undefined \def \showarticletitle #1{#1}   \fi
\ifx \showURL      \undefined \def \showURL       {\relax}        \fi
\providecommand\bibfield[2]{#2}
\providecommand\bibinfo[2]{#2}
\providecommand\natexlab[1]{#1}
\providecommand\showeprint[2][]{arXiv:#2}

\bibitem[Arbeláez et~al\mbox{.}(2011)]%
        {62}
\bibfield{author}{\bibinfo{person}{Pablo Arbeláez}, \bibinfo{person}{Michael
  Maire}, \bibinfo{person}{Charless Fowlkes}, {and} \bibinfo{person}{Jitendra
  Malik}.} \bibinfo{year}{2011}\natexlab{}.
\newblock \showarticletitle{Contour Detection and Hierarchical Image
  Segmentation}.
\newblock \bibinfo{journal}{\emph{IEEE Transactions on Pattern Analysis and
  Machine Intelligence}} \bibinfo{volume}{33}, \bibinfo{number}{5}
  (\bibinfo{year}{2011}), \bibinfo{pages}{898--916}.
\newblock


\bibitem[Cai et~al\mbox{.}(2016)]%
        {46}
\bibfield{author}{\bibinfo{person}{Bolun Cai}, \bibinfo{person}{Xiangmin Xu},
  \bibinfo{person}{Kui Jia}, \bibinfo{person}{Chunmei Qing}, {and}
  \bibinfo{person}{Dacheng Tao}.} \bibinfo{year}{2016}\natexlab{}.
\newblock \showarticletitle{DehazeNet: An End-to-End System for Single Image
  Haze Removal}.
\newblock \bibinfo{journal}{\emph{IEEE Transactions on Image Processing}}
  \bibinfo{volume}{25}, \bibinfo{number}{11} (\bibinfo{year}{2016}),
  \bibinfo{pages}{5187--5198}.
\newblock


\bibitem[Cai et~al\mbox{.}(2023)]%
        {27}
\bibfield{author}{\bibinfo{person}{Yuanhao Cai}, \bibinfo{person}{Hao Bian},
  \bibinfo{person}{Jing Lin}, \bibinfo{person}{Haoqian Wang},
  \bibinfo{person}{Radu Timofte}, {and} \bibinfo{person}{Yulun Zhang}.}
  \bibinfo{year}{2023}\natexlab{}.
\newblock \showarticletitle{Retinexformer: One-stage retinex-based transformer
  for low-light image enhancement}. In \bibinfo{booktitle}{\emph{Proceedings of
  the IEEE/CVF international conference on computer vision}}.
  \bibinfo{pages}{12504--12513}.
\newblock


\bibitem[Chan et~al\mbox{.}(2024)]%
        {29}
\bibfield{author}{\bibinfo{person}{Cheuk-Yiu Chan}, \bibinfo{person}{Wan-Chi
  Siu}, \bibinfo{person}{Yuk-Hee Chan}, {and} \bibinfo{person}{H.
  Anthony~Chan}.} \bibinfo{year}{2024}\natexlab{}.
\newblock \showarticletitle{AnlightenDiff: Anchoring Diffusion Probabilistic
  Model on Low Light Image Enhancement}.
\newblock \bibinfo{journal}{\emph{IEEE Transactions on Image Processing}}
  \bibinfo{volume}{33} (\bibinfo{year}{2024}), \bibinfo{pages}{6324--6339}.
\newblock


\bibitem[Chen and Li(2021)]%
        {17}
\bibfield{author}{\bibinfo{person}{Chenghao Chen} {and} \bibinfo{person}{Hao
  Li}.} \bibinfo{year}{2021}\natexlab{}.
\newblock \showarticletitle{Robust representation learning with feedback for
  single image deraining}. In \bibinfo{booktitle}{\emph{Proceedings of the
  IEEE/CVF conference on computer vision and pattern recognition}}.
  \bibinfo{pages}{7742--7751}.
\newblock


\bibitem[Chen et~al\mbox{.}(2025)]%
        {2}
\bibfield{author}{\bibinfo{person}{Jun Chen}, \bibinfo{person}{Liling Yang},
  \bibinfo{person}{Wei Yu}, \bibinfo{person}{Wenping Gong},
  \bibinfo{person}{Zhanchuan Cai}, {and} \bibinfo{person}{Jiayi Ma}.}
  \bibinfo{year}{2025}\natexlab{}.
\newblock \showarticletitle{SDSFusion: A Semantic-Aware Infrared and Visible
  Image Fusion Network for Degraded Scenes}.
\newblock \bibinfo{journal}{\emph{IEEE Transactions on Image Processing}}
  \bibinfo{volume}{34} (\bibinfo{year}{2025}), \bibinfo{pages}{3139--3153}.
\newblock


\bibitem[Chen et~al\mbox{.}(2022a)]%
        {69}
\bibfield{author}{\bibinfo{person}{Liangyu Chen}, \bibinfo{person}{Xiaojie
  Chu}, \bibinfo{person}{Xiangyu Zhang}, {and} \bibinfo{person}{Jian Sun}.}
  \bibinfo{year}{2022}\natexlab{a}.
\newblock \showarticletitle{Simple baselines for image restoration}. In
  \bibinfo{booktitle}{\emph{European conference on computer vision}}. Springer,
  \bibinfo{pages}{17--33}.
\newblock


\bibitem[Chen et~al\mbox{.}(2021)]%
        {51}
\bibfield{author}{\bibinfo{person}{Wei-Ting Chen}, \bibinfo{person}{Hao-Yu
  Fang}, \bibinfo{person}{Cheng-Lin Hsieh}, \bibinfo{person}{Cheng-Che Tsai},
  \bibinfo{person}{I Chen}, \bibinfo{person}{Jian-Jiun Ding},
  \bibinfo{person}{Sy-Yen Kuo}, {et~al\mbox{.}}}
  \bibinfo{year}{2021}\natexlab{}.
\newblock \showarticletitle{All snow removed: Single image desnowing algorithm
  using hierarchical dual-tree complex wavelet representation and contradict
  channel loss}. In \bibinfo{booktitle}{\emph{Proceedings of the IEEE/CVF
  international conference on computer vision}}. \bibinfo{pages}{4196--4205}.
\newblock


\bibitem[Chen et~al\mbox{.}(2022b)]%
        {58}
\bibfield{author}{\bibinfo{person}{Wei-Ting Chen}, \bibinfo{person}{Zhi-Kai
  Huang}, \bibinfo{person}{Cheng-Che Tsai}, \bibinfo{person}{Hao-Hsiang Yang},
  \bibinfo{person}{Jian-Jiun Ding}, {and} \bibinfo{person}{Sy-Yen Kuo}.}
  \bibinfo{year}{2022}\natexlab{b}.
\newblock \showarticletitle{Learning multiple adverse weather removal via
  two-stage knowledge learning and multi-contrastive regularization: Toward a
  unified model}. In \bibinfo{booktitle}{\emph{Proceedings of the IEEE/CVF
  conference on computer vision and pattern recognition}}.
  \bibinfo{pages}{17653--17662}.
\newblock


\bibitem[Cho et~al\mbox{.}(2021)]%
        {10}
\bibfield{author}{\bibinfo{person}{Sung-Jin Cho}, \bibinfo{person}{Seo-Won Ji},
  \bibinfo{person}{Jun-Pyo Hong}, \bibinfo{person}{Seung-Won Jung}, {and}
  \bibinfo{person}{Sung-Jea Ko}.} \bibinfo{year}{2021}\natexlab{}.
\newblock \showarticletitle{Rethinking coarse-to-fine approach in single image
  deblurring}. In \bibinfo{booktitle}{\emph{Proceedings of the IEEE/CVF
  international conference on computer vision}}. \bibinfo{pages}{4641--4650}.
\newblock


\bibitem[Conde et~al\mbox{.}(2024)]%
        {32}
\bibfield{author}{\bibinfo{person}{Marcos~V Conde}, \bibinfo{person}{Gregor
  Geigle}, {and} \bibinfo{person}{Radu Timofte}.}
  \bibinfo{year}{2024}\natexlab{}.
\newblock \showarticletitle{Instructir: High-quality image restoration
  following human instructions}. In \bibinfo{booktitle}{\emph{European
  Conference on Computer Vision}}. \bibinfo{pages}{1--21}.
\newblock


\bibitem[Cui et~al\mbox{.}(2023)]%
        {13}
\bibfield{author}{\bibinfo{person}{Yuning Cui}, \bibinfo{person}{Yi Tao},
  \bibinfo{person}{Wenqi Ren}, {and} \bibinfo{person}{Alois Knoll}.}
  \bibinfo{year}{2023}\natexlab{}.
\newblock \showarticletitle{Dual-domain attention for image deblurring}. In
  \bibinfo{booktitle}{\emph{Proceedings of the AAAI conference on artificial
  intelligence}}. \bibinfo{pages}{479--487}.
\newblock


\bibitem[Cui et~al\mbox{.}(2025)]%
        {59}
\bibfield{author}{\bibinfo{person}{Yuning Cui}, \bibinfo{person}{Syed~Waqas
  Zamir}, \bibinfo{person}{Salman Khan}, \bibinfo{person}{Alois Knoll},
  \bibinfo{person}{Mubarak Shah}, {and} \bibinfo{person}{Fahad~Shahbaz Khan}.}
  \bibinfo{year}{2025}\natexlab{}.
\newblock \showarticletitle{Adair: Adaptive all-in-one image restoration via
  frequency mining and modulation}. In \bibinfo{booktitle}{\emph{13th
  international conference on learning representations, ICLR 2025}}.
  \bibinfo{pages}{57335--57356}.
\newblock


\bibitem[Ding et~al\mbox{.}(2024)]%
        {8}
\bibfield{author}{\bibinfo{person}{Shifei Ding}, \bibinfo{person}{Qidong Wang},
  \bibinfo{person}{Lili Guo}, \bibinfo{person}{Xuan Li}, \bibinfo{person}{Ling
  Ding}, {and} \bibinfo{person}{Xindong Wu}.} \bibinfo{year}{2024}\natexlab{}.
\newblock \showarticletitle{Wavelet and Adaptive Coordinate Attention Guided
  Fine-Grained Residual Network for Image Denoising}.
\newblock \bibinfo{journal}{\emph{IEEE Transactions on Circuits and Systems for
  Video Technology}} \bibinfo{volume}{34}, \bibinfo{number}{7}
  (\bibinfo{year}{2024}), \bibinfo{pages}{6156--6166}.
\newblock


\bibitem[Dong et~al\mbox{.}(2020b)]%
        {47}
\bibfield{author}{\bibinfo{person}{Hang Dong}, \bibinfo{person}{Jinshan Pan},
  \bibinfo{person}{Lei Xiang}, \bibinfo{person}{Zhe Hu}, \bibinfo{person}{Xinyi
  Zhang}, \bibinfo{person}{Fei Wang}, {and} \bibinfo{person}{Ming-Hsuan Yang}.}
  \bibinfo{year}{2020}\natexlab{b}.
\newblock \showarticletitle{Multi-scale boosted dehazing network with dense
  feature fusion}. In \bibinfo{booktitle}{\emph{Proceedings of the IEEE/CVF
  conference on computer vision and pattern recognition}}.
  \bibinfo{pages}{2157--2167}.
\newblock


\bibitem[Dong et~al\mbox{.}(2020a)]%
        {su_3}
\bibfield{author}{\bibinfo{person}{Yu Dong}, \bibinfo{person}{Yihao Liu},
  \bibinfo{person}{He Zhang}, \bibinfo{person}{Shifeng Chen}, {and}
  \bibinfo{person}{Yu Qiao}.} \bibinfo{year}{2020}\natexlab{a}.
\newblock \showarticletitle{FD-GAN: Generative adversarial networks with
  fusion-discriminator for single image dehazing}. In
  \bibinfo{booktitle}{\emph{Proceedings of the AAAI conference on artificial
  intelligence}}. \bibinfo{pages}{10729--10736}.
\newblock


\bibitem[Gao et~al\mbox{.}(2019)]%
        {su_6}
\bibfield{author}{\bibinfo{person}{Hongyun Gao}, \bibinfo{person}{Xin Tao},
  \bibinfo{person}{Xiaoyong Shen}, {and} \bibinfo{person}{Jiaya Jia}.}
  \bibinfo{year}{2019}\natexlab{}.
\newblock \showarticletitle{Dynamic scene deblurring with parameter selective
  sharing and nested skip connections}. In
  \bibinfo{booktitle}{\emph{Proceedings of the IEEE/CVF conference on computer
  vision and pattern recognition}}. \bibinfo{pages}{3848--3856}.
\newblock


\bibitem[Guo et~al\mbox{.}(2020)]%
        {28}
\bibfield{author}{\bibinfo{person}{Chunle Guo}, \bibinfo{person}{Chongyi Li},
  \bibinfo{person}{Jichang Guo}, \bibinfo{person}{Chen~Change Loy},
  \bibinfo{person}{Junhui Hou}, \bibinfo{person}{Sam Kwong}, {and}
  \bibinfo{person}{Runmin Cong}.} \bibinfo{year}{2020}\natexlab{}.
\newblock \showarticletitle{Zero-reference deep curve estimation for low-light
  image enhancement}. In \bibinfo{booktitle}{\emph{Proceedings of the IEEE/CVF
  conference on computer vision and pattern recognition}}.
  \bibinfo{pages}{1780--1789}.
\newblock


\bibitem[Guo et~al\mbox{.}(2024)]%
        {67}
\bibfield{author}{\bibinfo{person}{Yu Guo}, \bibinfo{person}{Yuan Gao},
  \bibinfo{person}{Yuxu Lu}, \bibinfo{person}{Huilin Zhu},
  \bibinfo{person}{Ryan~Wen Liu}, {and} \bibinfo{person}{Shengfeng He}.}
  \bibinfo{year}{2024}\natexlab{}.
\newblock \showarticletitle{Onerestore: A universal restoration framework for
  composite degradation}. In \bibinfo{booktitle}{\emph{European conference on
  computer vision}}. \bibinfo{pages}{255--272}.
\newblock


\bibitem[He et~al\mbox{.}(2025)]%
        {25}
\bibfield{author}{\bibinfo{person}{Chunming He}, \bibinfo{person}{Chengyu
  Fang}, \bibinfo{person}{Yulun Zhang}, \bibinfo{person}{Longxiang Tang},
  \bibinfo{person}{Jinfa Huang}, \bibinfo{person}{Kai Li},
  \bibinfo{person}{zhenhua guo}, \bibinfo{person}{Xiu Li}, {and}
  \bibinfo{person}{Sina Farsiu}.} \bibinfo{year}{2025}\natexlab{}.
\newblock \showarticletitle{Reti-Diff: Illumination Degradation Image
  Restoration with Retinex-based Latent Diffusion Model}. In
  \bibinfo{booktitle}{\emph{International Conference on Learning
  Representations}}. \bibinfo{pages}{43332--43352}.
\newblock


\bibitem[Hu et~al\mbox{.}(2025)]%
        {76}
\bibfield{author}{\bibinfo{person}{Shengkai Hu}, \bibinfo{person}{Jiaqi Ma},
  \bibinfo{person}{Jun Wan}, \bibinfo{person}{Wenwen Min},
  \bibinfo{person}{Yongcheng Jing}, \bibinfo{person}{Lefei Zhang}, {and}
  \bibinfo{person}{Dacheng Tao}.} \bibinfo{year}{2025}\natexlab{}.
\newblock \showarticletitle{ClusIR: Towards Cluster-Guided All-in-One Image
  Restoration}.
\newblock \bibinfo{journal}{\emph{arXiv preprint arXiv:2512.10948}}
  (\bibinfo{year}{2025}).
\newblock


\bibitem[Huang et~al\mbox{.}(2021)]%
        {6}
\bibfield{author}{\bibinfo{person}{Tao Huang}, \bibinfo{person}{Songjiang Li},
  \bibinfo{person}{Xu Jia}, \bibinfo{person}{Huchuan Lu}, {and}
  \bibinfo{person}{Jianzhuang Liu}.} \bibinfo{year}{2021}\natexlab{}.
\newblock \showarticletitle{Neighbor2neighbor: Self-supervised denoising from
  single noisy images}. In \bibinfo{booktitle}{\emph{Proceedings of the
  IEEE/CVF conference on computer vision and pattern recognition}}.
  \bibinfo{pages}{14781--14790}.
\newblock


\bibitem[Jiang et~al\mbox{.}(2020)]%
        {15}
\bibfield{author}{\bibinfo{person}{Kui Jiang}, \bibinfo{person}{Zhongyuan
  Wang}, \bibinfo{person}{Peng Yi}, \bibinfo{person}{Chen Chen},
  \bibinfo{person}{Baojin Huang}, \bibinfo{person}{Yimin Luo},
  \bibinfo{person}{Jiayi Ma}, {and} \bibinfo{person}{Junjun Jiang}.}
  \bibinfo{year}{2020}\natexlab{}.
\newblock \showarticletitle{Multi-scale progressive fusion network for single
  image deraining}. In \bibinfo{booktitle}{\emph{Proceedings of the IEEE/CVF
  conference on computer vision and pattern recognition}}.
  \bibinfo{pages}{8346--8355}.
\newblock


\bibitem[Kingma and Ba(2014)]%
        {73}
\bibfield{author}{\bibinfo{person}{Diederik~P Kingma} {and}
  \bibinfo{person}{Jimmy Ba}.} \bibinfo{year}{2014}\natexlab{}.
\newblock \showarticletitle{Adam: A method for stochastic optimization}.
\newblock \bibinfo{journal}{\emph{arXiv preprint arXiv:1412.6980}}
  (\bibinfo{year}{2014}).
\newblock


\bibitem[Kupyn et~al\mbox{.}(2019)]%
        {11}
\bibfield{author}{\bibinfo{person}{Orest Kupyn}, \bibinfo{person}{Tetiana
  Martyniuk}, \bibinfo{person}{Junru Wu}, {and} \bibinfo{person}{Zhangyang
  Wang}.} \bibinfo{year}{2019}\natexlab{}.
\newblock \showarticletitle{Deblurgan-v2: Deblurring (orders-of-magnitude)
  faster and better}. In \bibinfo{booktitle}{\emph{Proceedings of the IEEE/CVF
  international conference on computer vision}}. \bibinfo{pages}{8878--8887}.
\newblock


\bibitem[Li et~al\mbox{.}(2022)]%
        {33}
\bibfield{author}{\bibinfo{person}{Boyun Li}, \bibinfo{person}{Xiao Liu},
  \bibinfo{person}{Peng Hu}, \bibinfo{person}{Zhongqin Wu},
  \bibinfo{person}{Jiancheng Lv}, {and} \bibinfo{person}{Xi Peng}.}
  \bibinfo{year}{2022}\natexlab{}.
\newblock \showarticletitle{All-in-one image restoration for unknown
  corruption}. In \bibinfo{booktitle}{\emph{Proceedings of the IEEE/CVF
  conference on computer vision and pattern recognition}}.
  \bibinfo{pages}{17452--17462}.
\newblock


\bibitem[Li et~al\mbox{.}(2019)]%
        {60}
\bibfield{author}{\bibinfo{person}{Boyi Li}, \bibinfo{person}{Wenqi Ren},
  \bibinfo{person}{Dengpan Fu}, \bibinfo{person}{Dacheng Tao},
  \bibinfo{person}{Dan Feng}, \bibinfo{person}{Wenjun Zeng}, {and}
  \bibinfo{person}{Zhangyang Wang}.} \bibinfo{year}{2019}\natexlab{}.
\newblock \showarticletitle{Benchmarking Single-Image Dehazing and Beyond}.
\newblock \bibinfo{journal}{\emph{IEEE Transactions on Image Processing}}
  \bibinfo{volume}{28}, \bibinfo{number}{1} (\bibinfo{year}{2019}),
  \bibinfo{pages}{492--505}.
\newblock


\bibitem[Li et~al\mbox{.}(2020)]%
        {57}
\bibfield{author}{\bibinfo{person}{Ruoteng Li}, \bibinfo{person}{Robby~T Tan},
  {and} \bibinfo{person}{Loong-Fah Cheong}.} \bibinfo{year}{2020}\natexlab{}.
\newblock \showarticletitle{All in one bad weather removal using architectural
  search}. In \bibinfo{booktitle}{\emph{Proceedings of the IEEE/CVF conference
  on computer vision and pattern recognition}}. \bibinfo{pages}{3175--3185}.
\newblock


\bibitem[Li et~al\mbox{.}(2023)]%
        {34}
\bibfield{author}{\bibinfo{person}{Zilong Li}, \bibinfo{person}{Yiming Lei},
  \bibinfo{person}{Chenglong Ma}, \bibinfo{person}{Junping Zhang}, {and}
  \bibinfo{person}{Hongming Shan}.} \bibinfo{year}{2023}\natexlab{}.
\newblock \showarticletitle{Prompt-In-Prompt Learning for Universal Image
  Restoration}.
\newblock \bibinfo{journal}{\emph{arXiv preprint arXiv:2312.05038}}
  (\bibinfo{year}{2023}).
\newblock


\bibitem[Liang et~al\mbox{.}(2021)]%
        {56}
\bibfield{author}{\bibinfo{person}{Jingyun Liang}, \bibinfo{person}{Jiezhang
  Cao}, \bibinfo{person}{Guolei Sun}, \bibinfo{person}{Kai Zhang},
  \bibinfo{person}{Luc Van~Gool}, {and} \bibinfo{person}{Radu Timofte}.}
  \bibinfo{year}{2021}\natexlab{}.
\newblock \showarticletitle{Swinir: Image restoration using swin transformer}.
  In \bibinfo{booktitle}{\emph{Proceedings of the IEEE/CVF international
  conference on computer vision}}. \bibinfo{pages}{1833--1844}.
\newblock


\bibitem[Lin et~al\mbox{.}(2023)]%
        {7}
\bibfield{author}{\bibinfo{person}{Xin Lin}, \bibinfo{person}{Chao Ren},
  \bibinfo{person}{Xiao Liu}, \bibinfo{person}{Jie Huang}, {and}
  \bibinfo{person}{Yinjie Lei}.} \bibinfo{year}{2023}\natexlab{}.
\newblock \showarticletitle{Unsupervised image denoising in real-world
  scenarios via self-collaboration parallel generative adversarial branches}.
  In \bibinfo{booktitle}{\emph{Proceedings of the IEEE/CVF International
  Conference on Computer Vision}}. \bibinfo{pages}{12642--12652}.
\newblock


\bibitem[Liu et~al\mbox{.}(2026)]%
        {23}
\bibfield{author}{\bibinfo{person}{Yun Liu}, \bibinfo{person}{Tao Li},
  \bibinfo{person}{Chunping Tan}, \bibinfo{person}{Wenqi Ren},
  \bibinfo{person}{Cosmin Ancuti}, {and} \bibinfo{person}{Weisi Lin}.}
  \bibinfo{year}{2026}\natexlab{}.
\newblock \showarticletitle{IHDCP: Single Image Dehazing Using Inverted Haze
  Density Correction Prior}.
\newblock \bibinfo{journal}{\emph{IEEE Transactions on Image Processing}}
  \bibinfo{volume}{35} (\bibinfo{year}{2026}), \bibinfo{pages}{1448--1461}.
\newblock


\bibitem[Liu et~al\mbox{.}(2024)]%
        {14}
\bibfield{author}{\bibinfo{person}{Zhaoxin Liu}, \bibinfo{person}{Jinjian Wu},
  \bibinfo{person}{Guangming Shi}, \bibinfo{person}{Wen Yang},
  \bibinfo{person}{Weisheng Dong}, {and} \bibinfo{person}{Qinghang Zhao}.}
  \bibinfo{year}{2024}\natexlab{}.
\newblock \showarticletitle{Motion-Oriented Hybrid Spiking Neural Networks for
  Event-Based Motion Deblurring}.
\newblock \bibinfo{journal}{\emph{IEEE Transactions on Circuits and Systems for
  Video Technology}} \bibinfo{volume}{34}, \bibinfo{number}{5}
  (\bibinfo{year}{2024}), \bibinfo{pages}{3742--3754}.
\newblock


\bibitem[Luo et~al\mbox{.}(2024)]%
        {35}
\bibfield{author}{\bibinfo{person}{Ziwei Luo}, \bibinfo{person}{Fredrik~K.
  Gustafsson}, \bibinfo{person}{Zheng Zhao}, \bibinfo{person}{Jens
  Sj\"{o}lund}, {and} \bibinfo{person}{Thomas Sch\"{o}n}.}
  \bibinfo{year}{2024}\natexlab{}.
\newblock \showarticletitle{Controlling Vision-Language Models for Multi-Task
  Image Restoration}. In \bibinfo{booktitle}{\emph{International Conference on
  Learning Representations}}. \bibinfo{pages}{16226--16246}.
\newblock


\bibitem[Ma et~al\mbox{.}(2025)]%
        {77}
\bibfield{author}{\bibinfo{person}{Jiaqi Ma}, \bibinfo{person}{Shengkai Hu},
  \bibinfo{person}{Xu Zhang}, \bibinfo{person}{Jun Wan},
  \bibinfo{person}{Jiaxing Huang}, \bibinfo{person}{Lefei Zhang}, {and}
  \bibinfo{person}{Salman Khan}.} \bibinfo{year}{2025}\natexlab{}.
\newblock \showarticletitle{EvoIR: Towards All-in-One Image Restoration via
  Evolutionary Frequency Modulation}.
\newblock \bibinfo{journal}{\emph{arXiv preprint arXiv:2512.05104}}
  (\bibinfo{year}{2025}).
\newblock


\bibitem[Ma et~al\mbox{.}(2017)]%
        {63}
\bibfield{author}{\bibinfo{person}{Kede Ma}, \bibinfo{person}{Zhengfang
  Duanmu}, \bibinfo{person}{Qingbo Wu}, \bibinfo{person}{Zhou Wang},
  \bibinfo{person}{Hongwei Yong}, \bibinfo{person}{Hongliang Li}, {and}
  \bibinfo{person}{Lei Zhang}.} \bibinfo{year}{2017}\natexlab{}.
\newblock \showarticletitle{Waterloo Exploration Database: New Challenges for
  Image Quality Assessment Models}.
\newblock \bibinfo{journal}{\emph{IEEE Transactions on Image Processing}}
  \bibinfo{volume}{26}, \bibinfo{number}{2} (\bibinfo{year}{2017}),
  \bibinfo{pages}{1004--1016}.
\newblock


\bibitem[Mao et~al\mbox{.}(2023)]%
        {50}
\bibfield{author}{\bibinfo{person}{Xintian Mao}, \bibinfo{person}{Yiming Liu},
  \bibinfo{person}{Fengze Liu}, \bibinfo{person}{Qingli Li},
  \bibinfo{person}{Wei Shen}, {and} \bibinfo{person}{Yan Wang}.}
  \bibinfo{year}{2023}\natexlab{}.
\newblock \showarticletitle{Intriguing findings of frequency selection for
  image deblurring}. In \bibinfo{booktitle}{\emph{Proceedings of the AAAI
  conference on artificial intelligence}}. \bibinfo{pages}{1905--1913}.
\newblock


\bibitem[Martin et~al\mbox{.}(2001)]%
        {64}
\bibfield{author}{\bibinfo{person}{D. Martin}, \bibinfo{person}{C. Fowlkes},
  \bibinfo{person}{D. Tal}, {and} \bibinfo{person}{J. Malik}.}
  \bibinfo{year}{2001}\natexlab{}.
\newblock \showarticletitle{A database of human segmented natural images and
  its application to evaluating segmentation algorithms and measuring
  ecological statistics}. In \bibinfo{booktitle}{\emph{Proceedings Eighth IEEE
  International Conference on Computer Vision. ICCV 2001}},
  Vol.~\bibinfo{volume}{2}. \bibinfo{pages}{416--423 vol.2}.
\newblock


\bibitem[Nah et~al\mbox{.}(2017)]%
        {65}
\bibfield{author}{\bibinfo{person}{Seungjun Nah}, \bibinfo{person}{Tae
  Hyun~Kim}, {and} \bibinfo{person}{Kyoung Mu~Lee}.}
  \bibinfo{year}{2017}\natexlab{}.
\newblock \showarticletitle{Deep multi-scale convolutional neural network for
  dynamic scene deblurring}. In \bibinfo{booktitle}{\emph{Proceedings of the
  IEEE conference on computer vision and pattern recognition}}.
  \bibinfo{pages}{3883--3891}.
\newblock


\bibitem[Park et~al\mbox{.}(2020)]%
        {12}
\bibfield{author}{\bibinfo{person}{Dongwon Park}, \bibinfo{person}{Dong~Un
  Kang}, \bibinfo{person}{Jisoo Kim}, {and} \bibinfo{person}{Se~Young Chun}.}
  \bibinfo{year}{2020}\natexlab{}.
\newblock \showarticletitle{Multi-temporal recurrent neural networks for
  progressive non-uniform single image deblurring with incremental temporal
  training}. In \bibinfo{booktitle}{\emph{European conference on computer
  vision}}. \bibinfo{pages}{327--343}.
\newblock


\bibitem[Potlapalli et~al\mbox{.}(2023)]%
        {36}
\bibfield{author}{\bibinfo{person}{Vaishnav Potlapalli},
  \bibinfo{person}{Syed~Waqas Zamir}, \bibinfo{person}{Salman~H Khan}, {and}
  \bibinfo{person}{Fahad Shahbaz~Khan}.} \bibinfo{year}{2023}\natexlab{}.
\newblock \showarticletitle{Promptir: Prompting for all-in-one image
  restoration}.
\newblock \bibinfo{journal}{\emph{Advances in neural information processing
  systems}}  \bibinfo{volume}{36} (\bibinfo{year}{2023}),
  \bibinfo{pages}{71275--71293}.
\newblock


\bibitem[Qin et~al\mbox{.}(2020)]%
        {21}
\bibfield{author}{\bibinfo{person}{Xu Qin}, \bibinfo{person}{Zhilin Wang},
  \bibinfo{person}{Yuanchao Bai}, \bibinfo{person}{Xiaodong Xie}, {and}
  \bibinfo{person}{Huizhu Jia}.} \bibinfo{year}{2020}\natexlab{}.
\newblock \showarticletitle{FFA-Net: Feature fusion attention network for
  single image dehazing}. In \bibinfo{booktitle}{\emph{Proceedings of the AAAI
  conference on artificial intelligence}}. \bibinfo{pages}{11908--11915}.
\newblock


\bibitem[Qu et~al\mbox{.}(2019)]%
        {su_2}
\bibfield{author}{\bibinfo{person}{Yanyun Qu}, \bibinfo{person}{Yizi Chen},
  \bibinfo{person}{Jingying Huang}, {and} \bibinfo{person}{Yuan Xie}.}
  \bibinfo{year}{2019}\natexlab{}.
\newblock \showarticletitle{Enhanced pix2pix dehazing network}. In
  \bibinfo{booktitle}{\emph{Proceedings of the IEEE/CVF conference on computer
  vision and pattern recognition}}. \bibinfo{pages}{8160--8168}.
\newblock


\bibitem[Ren et~al\mbox{.}(2021)]%
        {9}
\bibfield{author}{\bibinfo{person}{Chao Ren}, \bibinfo{person}{Xiaohai He},
  \bibinfo{person}{Chuncheng Wang}, {and} \bibinfo{person}{Zhibo Zhao}.}
  \bibinfo{year}{2021}\natexlab{}.
\newblock \showarticletitle{Adaptive consistency prior based deep network for
  image denoising}. In \bibinfo{booktitle}{\emph{Proceedings of the IEEE/CVF
  conference on computer vision and pattern recognition}}.
  \bibinfo{pages}{8596--8606}.
\newblock


\bibitem[Ren et~al\mbox{.}(2019)]%
        {16}
\bibfield{author}{\bibinfo{person}{Dongwei Ren}, \bibinfo{person}{Wangmeng
  Zuo}, \bibinfo{person}{Qinghua Hu}, \bibinfo{person}{Pengfei Zhu}, {and}
  \bibinfo{person}{Deyu Meng}.} \bibinfo{year}{2019}\natexlab{}.
\newblock \showarticletitle{Progressive image deraining networks: A better and
  simpler baseline}. In \bibinfo{booktitle}{\emph{Proceedings of the IEEE/CVF
  conference on computer vision and pattern recognition}}.
  \bibinfo{pages}{3937--3946}.
\newblock


\bibitem[Ren et~al\mbox{.}(2016)]%
        {su_1}
\bibfield{author}{\bibinfo{person}{Wenqi Ren}, \bibinfo{person}{Si Liu},
  \bibinfo{person}{Hua Zhang}, \bibinfo{person}{Jinshan Pan},
  \bibinfo{person}{Xiaochun Cao}, {and} \bibinfo{person}{Ming-Hsuan Yang}.}
  \bibinfo{year}{2016}\natexlab{}.
\newblock \showarticletitle{Single image dehazing via multi-scale convolutional
  neural networks}. In \bibinfo{booktitle}{\emph{European conference on
  computer vision}}. \bibinfo{pages}{154--169}.
\newblock


\bibitem[Ruan et~al\mbox{.}(2022)]%
        {49}
\bibfield{author}{\bibinfo{person}{Lingyan Ruan}, \bibinfo{person}{Bin Chen},
  \bibinfo{person}{Jizhou Li}, {and} \bibinfo{person}{Miuling Lam}.}
  \bibinfo{year}{2022}\natexlab{}.
\newblock \showarticletitle{Learning to deblur using light field generated and
  real defocus images}. In \bibinfo{booktitle}{\emph{Proceedings of the
  IEEE/CVF conference on computer vision and pattern recognition}}.
  \bibinfo{pages}{16304--16313}.
\newblock


\bibitem[Su et~al\mbox{.}(2025)]%
        {52}
\bibfield{author}{\bibinfo{person}{Xiongfei Su}, \bibinfo{person}{Siyuan Li},
  \bibinfo{person}{Yuning Cui}, \bibinfo{person}{Miao Cao},
  \bibinfo{person}{Yulun Zhang}, \bibinfo{person}{Zheng Chen},
  \bibinfo{person}{Zongliang Wu}, \bibinfo{person}{Zedong Wang},
  \bibinfo{person}{Yuanlong Zhang}, {and} \bibinfo{person}{Xin Yuan}.}
  \bibinfo{year}{2025}\natexlab{}.
\newblock \showarticletitle{Prior-guided hierarchical harmonization network for
  efficient image dehazing}. In \bibinfo{booktitle}{\emph{Proceedings of the
  AAAI conference on artificial intelligence}}. \bibinfo{pages}{7042--7050}.
\newblock


\bibitem[Tian et~al\mbox{.}(2020)]%
        {su_8}
\bibfield{author}{\bibinfo{person}{Chunwei Tian}, \bibinfo{person}{Yong Xu},
  {and} \bibinfo{person}{Wangmeng Zuo}.} \bibinfo{year}{2020}\natexlab{}.
\newblock \showarticletitle{Image denoising using deep CNN with batch
  renormalization}.
\newblock \bibinfo{journal}{\emph{Neural Networks}}  \bibinfo{volume}{121}
  (\bibinfo{year}{2020}), \bibinfo{pages}{461--473}.
\newblock


\bibitem[Tian et~al\mbox{.}(2025)]%
        {42}
\bibfield{author}{\bibinfo{person}{Xiangpeng Tian}, \bibinfo{person}{Xiangyu
  Liao}, \bibinfo{person}{Xiao Liu}, \bibinfo{person}{Meng Li}, {and}
  \bibinfo{person}{Chao Ren}.} \bibinfo{year}{2025}\natexlab{}.
\newblock \showarticletitle{Degradation-aware feature perturbation for
  all-in-one image restoration}. In \bibinfo{booktitle}{\emph{Proceedings of
  the IEEE/CVF Conference on Computer Vision and Pattern Recognition}}.
  \bibinfo{pages}{28165--28175}.
\newblock


\bibitem[Tian et~al\mbox{.}(2024)]%
        {37}
\bibfield{author}{\bibinfo{person}{Yuchuan Tian}, \bibinfo{person}{Jianhong
  Han}, \bibinfo{person}{Hanting Chen}, \bibinfo{person}{Yuanyuan Xi},
  \bibinfo{person}{Ning Ding}, \bibinfo{person}{Jie Hu}, \bibinfo{person}{Chao
  Xu}, {and} \bibinfo{person}{Yunhe Wang}.} \bibinfo{year}{2024}\natexlab{}.
\newblock \showarticletitle{Instruct-ipt: All-in-one image processing
  transformer via weight modulation}.
\newblock \bibinfo{journal}{\emph{arXiv preprint arXiv:2407.00676}}
  (\bibinfo{year}{2024}).
\newblock


\bibitem[Valanarasu et~al\mbox{.}(2022)]%
        {38}
\bibfield{author}{\bibinfo{person}{Jeya Maria~Jose Valanarasu},
  \bibinfo{person}{Rajeev Yasarla}, {and} \bibinfo{person}{Vishal~M Patel}.}
  \bibinfo{year}{2022}\natexlab{}.
\newblock \showarticletitle{Transweather: Transformer-based restoration of
  images degraded by adverse weather conditions}. In
  \bibinfo{booktitle}{\emph{Proceedings of the IEEE/CVF conference on computer
  vision and pattern recognition}}. \bibinfo{pages}{2353--2363}.
\newblock


\bibitem[Vaswani et~al\mbox{.}(2017)]%
        {54}
\bibfield{author}{\bibinfo{person}{Ashish Vaswani}, \bibinfo{person}{Noam
  Shazeer}, \bibinfo{person}{Niki Parmar}, \bibinfo{person}{Jakob Uszkoreit},
  \bibinfo{person}{Llion Jones}, \bibinfo{person}{Aidan~N Gomez},
  \bibinfo{person}{{\L}ukasz Kaiser}, {and} \bibinfo{person}{Illia
  Polosukhin}.} \bibinfo{year}{2017}\natexlab{}.
\newblock \showarticletitle{Attention is all you need}.
\newblock \bibinfo{journal}{\emph{Advances in neural information processing
  systems}}  \bibinfo{volume}{30} (\bibinfo{year}{2017}).
\newblock


\bibitem[Wang et~al\mbox{.}(2024)]%
        {24}
\bibfield{author}{\bibinfo{person}{Cong Wang}, \bibinfo{person}{Jinshan Pan},
  \bibinfo{person}{Wanyu Lin}, \bibinfo{person}{Jiangxin Dong},
  \bibinfo{person}{Wei Wang}, {and} \bibinfo{person}{Xiao-Ming Wu}.}
  \bibinfo{year}{2024}\natexlab{}.
\newblock \showarticletitle{Selfpromer: Self-prompt dehazing transformers with
  depth-consistency}. In \bibinfo{booktitle}{\emph{Proceedings of the AAAI
  conference on artificial intelligence}}, Vol.~\bibinfo{volume}{38}.
  \bibinfo{pages}{5327--5335}.
\newblock


\bibitem[Wang et~al\mbox{.}(2025)]%
        {79}
\bibfield{author}{\bibinfo{person}{Yongzhen Wang}, \bibinfo{person}{Yongjun
  Li}, \bibinfo{person}{Zhuoran Zheng}, \bibinfo{person}{Xiao-Ping Zhang},
  {and} \bibinfo{person}{Mingqiang Wei}.} \bibinfo{year}{2025}\natexlab{}.
\newblock \showarticletitle{M2Restore: Mixture-of-Experts-Based Mamba-CNN
  Fusion Framework for All-in-One Image Restoration}.
\newblock \bibinfo{journal}{\emph{IEEE Transactions on Image Processing}}
  \bibinfo{volume}{34} (\bibinfo{year}{2025}), \bibinfo{pages}{8086--8100}.
\newblock


\bibitem[Wang et~al\mbox{.}(2022)]%
        {30}
\bibfield{author}{\bibinfo{person}{Zhendong Wang}, \bibinfo{person}{Xiaodong
  Cun}, \bibinfo{person}{Jianmin Bao}, \bibinfo{person}{Wengang Zhou},
  \bibinfo{person}{Jianzhuang Liu}, {and} \bibinfo{person}{Houqiang Li}.}
  \bibinfo{year}{2022}\natexlab{}.
\newblock \showarticletitle{Uformer: A general u-shaped transformer for image
  restoration}. In \bibinfo{booktitle}{\emph{Proceedings of the IEEE/CVF
  conference on computer vision and pattern recognition}}.
  \bibinfo{pages}{17683--17693}.
\newblock


\bibitem[Wei et~al\mbox{.}(2018)]%
        {66}
\bibfield{author}{\bibinfo{person}{Chen Wei}, \bibinfo{person}{Wenjing Wang},
  \bibinfo{person}{Wenhan Yang}, {and} \bibinfo{person}{Jiaying Liu}.}
  \bibinfo{year}{2018}\natexlab{}.
\newblock \showarticletitle{Deep retinex decomposition for low-light
  enhancement}.
\newblock \bibinfo{journal}{\emph{arXiv preprint arXiv:1808.04560}}
  (\bibinfo{year}{2018}).
\newblock


\bibitem[Wei et~al\mbox{.}(2019)]%
        {su_5}
\bibfield{author}{\bibinfo{person}{Wei Wei}, \bibinfo{person}{Deyu Meng},
  \bibinfo{person}{Qian Zhao}, \bibinfo{person}{Zongben Xu}, {and}
  \bibinfo{person}{Ying Wu}.} \bibinfo{year}{2019}\natexlab{}.
\newblock \showarticletitle{Semi-supervised transfer learning for image rain
  removal}. In \bibinfo{booktitle}{\emph{Proceedings of the IEEE/CVF conference
  on computer vision and pattern recognition}}. \bibinfo{pages}{3877--3886}.
\newblock


\bibitem[Wu et~al\mbox{.}(2026)]%
        {74}
\bibfield{author}{\bibinfo{person}{Gang Wu}, \bibinfo{person}{Junjun Jiang},
  \bibinfo{person}{Kui Jiang}, \bibinfo{person}{Xianming Liu}, {and}
  \bibinfo{person}{Liqiang Nie}.} \bibinfo{year}{2026}\natexlab{}.
\newblock \showarticletitle{Beyond Degradation Redundancy: Contrastive Prompt
  Learning for All-in-One Image Restoration}.
\newblock \bibinfo{journal}{\emph{IEEE Transactions on Pattern Analysis and
  Machine Intelligence}} \bibinfo{volume}{48}, \bibinfo{number}{4}
  (\bibinfo{year}{2026}), \bibinfo{pages}{4005--4022}.
\newblock


\bibitem[Wu et~al\mbox{.}(2025a)]%
        {75}
\bibfield{author}{\bibinfo{person}{Gang Wu}, \bibinfo{person}{Junjun Jiang},
  \bibinfo{person}{Yijun Wang}, \bibinfo{person}{Kui Jiang}, {and}
  \bibinfo{person}{Xianming Liu}.} \bibinfo{year}{2025}\natexlab{a}.
\newblock \showarticletitle{Debiased all-in-one image restoration with task
  uncertainty regularization}. In \bibinfo{booktitle}{\emph{Proceedings of the
  AAAI Conference on Artificial Intelligence}}, Vol.~\bibinfo{volume}{39}.
  \bibinfo{pages}{8386--8394}.
\newblock


\bibitem[Wu et~al\mbox{.}(2021)]%
        {22}
\bibfield{author}{\bibinfo{person}{Haiyan Wu}, \bibinfo{person}{Yanyun Qu},
  \bibinfo{person}{Shaohui Lin}, \bibinfo{person}{Jian Zhou},
  \bibinfo{person}{Ruizhi Qiao}, \bibinfo{person}{Zhizhong Zhang},
  \bibinfo{person}{Yuan Xie}, {and} \bibinfo{person}{Lizhuang Ma}.}
  \bibinfo{year}{2021}\natexlab{}.
\newblock \showarticletitle{Contrastive learning for compact single image
  dehazing}. In \bibinfo{booktitle}{\emph{Proceedings of the IEEE/CVF
  conference on computer vision and pattern recognition}}.
  \bibinfo{pages}{10551--10560}.
\newblock


\bibitem[Wu et~al\mbox{.}(2025b)]%
        {45}
\bibfield{author}{\bibinfo{person}{Jiawei Wu}, \bibinfo{person}{Zhifei Yang},
  \bibinfo{person}{Zhe Wang}, {and} \bibinfo{person}{Zhi Jin}.}
  \bibinfo{year}{2025}\natexlab{b}.
\newblock \showarticletitle{Gradient as conditions: Rethinking HOG for
  all-in-one image restoration}.
\newblock \bibinfo{journal}{\emph{arXiv preprint arXiv:2504.09377}}
  (\bibinfo{year}{2025}).
\newblock


\bibitem[Xu et~al\mbox{.}(2025)]%
        {1}
\bibfield{author}{\bibinfo{person}{Han Xu}, \bibinfo{person}{Xunpeng Yi},
  \bibinfo{person}{Chen Lu}, \bibinfo{person}{Guangcan Liu}, {and}
  \bibinfo{person}{Jiayi Ma}.} \bibinfo{year}{2025}\natexlab{}.
\newblock \showarticletitle{URFusion: Unsupervised Unified Degradation-Robust
  Image Fusion Network}.
\newblock \bibinfo{journal}{\emph{IEEE Transactions on Image Processing}}
  \bibinfo{volume}{34} (\bibinfo{year}{2025}), \bibinfo{pages}{5803--5818}.
\newblock


\bibitem[Yan et~al\mbox{.}(2025)]%
        {26}
\bibfield{author}{\bibinfo{person}{Qingsen Yan}, \bibinfo{person}{Yixu Feng},
  \bibinfo{person}{Cheng Zhang}, \bibinfo{person}{Guansong Pang},
  \bibinfo{person}{Kangbiao Shi}, \bibinfo{person}{Peng Wu},
  \bibinfo{person}{Wei Dong}, \bibinfo{person}{Jinqiu Sun}, {and}
  \bibinfo{person}{Yanning Zhang}.} \bibinfo{year}{2025}\natexlab{}.
\newblock \showarticletitle{Hvi: A new color space for low-light image
  enhancement}. In \bibinfo{booktitle}{\emph{Proceedings of the computer vision
  and pattern recognition conference}}. \bibinfo{pages}{5678--5687}.
\newblock


\bibitem[Yang et~al\mbox{.}(2020a)]%
        {61}
\bibfield{author}{\bibinfo{person}{Wenhan Yang}, \bibinfo{person}{Robby~T.
  Tan}, \bibinfo{person}{Jiashi Feng}, \bibinfo{person}{Zongming Guo},
  \bibinfo{person}{Shuicheng Yan}, {and} \bibinfo{person}{Jiaying Liu}.}
  \bibinfo{year}{2020}\natexlab{a}.
\newblock \showarticletitle{Joint Rain Detection and Removal from a Single
  Image with Contextualized Deep Networks}.
\newblock \bibinfo{journal}{\emph{IEEE Transactions on Pattern Analysis and
  Machine Intelligence}} \bibinfo{volume}{42}, \bibinfo{number}{6}
  (\bibinfo{year}{2020}), \bibinfo{pages}{1377--1393}.
\newblock


\bibitem[Yang et~al\mbox{.}(2020b)]%
        {19}
\bibfield{author}{\bibinfo{person}{Wenhan Yang}, \bibinfo{person}{Robby~T Tan},
  \bibinfo{person}{Shiqi Wang}, \bibinfo{person}{Yuming Fang}, {and}
  \bibinfo{person}{Jiaying Liu}.} \bibinfo{year}{2020}\natexlab{b}.
\newblock \showarticletitle{Single image deraining: From model-based to
  data-driven and beyond}.
\newblock \bibinfo{journal}{\emph{IEEE Transactions on pattern analysis and
  machine intelligence}} \bibinfo{volume}{43}, \bibinfo{number}{11}
  (\bibinfo{year}{2020}), \bibinfo{pages}{4059--4077}.
\newblock


\bibitem[Yao et~al\mbox{.}(2024)]%
        {re_1}
\bibfield{author}{\bibinfo{person}{Mingde Yao}, \bibinfo{person}{Ruikang Xu},
  \bibinfo{person}{Yuanshen Guan}, \bibinfo{person}{Jie Huang}, {and}
  \bibinfo{person}{Zhiwei Xiong}.} \bibinfo{year}{2024}\natexlab{}.
\newblock \showarticletitle{Neural Degradation Representation Learning for
  All-in-One Image Restoration}.
\newblock \bibinfo{journal}{\emph{IEEE Transactions on Image Processing}}
  \bibinfo{volume}{33} (\bibinfo{year}{2024}), \bibinfo{pages}{5408--5423}.
\newblock


\bibitem[Yao et~al\mbox{.}(2025)]%
        {4}
\bibfield{author}{\bibinfo{person}{Siyuan Yao}, \bibinfo{person}{Rui Zhu},
  \bibinfo{person}{Ziqi Wang}, \bibinfo{person}{Wenqi Ren},
  \bibinfo{person}{Yanyang Yan}, {and} \bibinfo{person}{Xiaochun Cao}.}
  \bibinfo{year}{2025}\natexlab{}.
\newblock \showarticletitle{UMDATrack: Unified Multi-Domain Adaptive Tracking
  Under Adverse Weather Conditions}. In \bibinfo{booktitle}{\emph{Proceedings
  of the IEEE/CVF International Conference on Computer Vision (ICCV)}}.
  \bibinfo{pages}{6466--6475}.
\newblock


\bibitem[Yasarla and Patel(2019)]%
        {su_4}
\bibfield{author}{\bibinfo{person}{Rajeev Yasarla} {and}
  \bibinfo{person}{Vishal~M Patel}.} \bibinfo{year}{2019}\natexlab{}.
\newblock \showarticletitle{Uncertainty guided multi-scale residual
  learning-using a cycle spinning cnn for single image de-raining}. In
  \bibinfo{booktitle}{\emph{Proceedings of the IEEE/CVF conference on computer
  vision and pattern recognition}}. \bibinfo{pages}{8405--8414}.
\newblock


\bibitem[Yu et~al\mbox{.}(2025)]%
        {43}
\bibfield{author}{\bibinfo{person}{Xiaoyan Yu}, \bibinfo{person}{Shen Zhou},
  \bibinfo{person}{Huafeng Li}, {and} \bibinfo{person}{Liehuang Zhu}.}
  \bibinfo{year}{2025}\natexlab{}.
\newblock \showarticletitle{Multi-Expert Adaptive Selection: Task-Balancing for
  All-in-One Image Restoration}.
\newblock \bibinfo{journal}{\emph{IEEE Transactions on Circuits and Systems for
  Video Technology}} \bibinfo{volume}{35}, \bibinfo{number}{5}
  (\bibinfo{year}{2025}), \bibinfo{pages}{4619--4634}.
\newblock


\bibitem[Zamfir et~al\mbox{.}(2025)]%
        {44}
\bibfield{author}{\bibinfo{person}{Eduard Zamfir}, \bibinfo{person}{Zongwei
  Wu}, \bibinfo{person}{Nancy Mehta}, \bibinfo{person}{Yuedong Tan},
  \bibinfo{person}{Danda~Pani Paudel}, \bibinfo{person}{Yulun Zhang}, {and}
  \bibinfo{person}{Radu Timofte}.} \bibinfo{year}{2025}\natexlab{}.
\newblock \showarticletitle{Complexity experts are task-discriminative learners
  for any image restoration}. In \bibinfo{booktitle}{\emph{Proceedings of the
  Computer Vision and Pattern Recognition Conference}}.
  \bibinfo{pages}{12753--12763}.
\newblock


\bibitem[Zamir et~al\mbox{.}(2022)]%
        {31}
\bibfield{author}{\bibinfo{person}{Syed~Waqas Zamir}, \bibinfo{person}{Aditya
  Arora}, \bibinfo{person}{Salman Khan}, \bibinfo{person}{Munawar Hayat},
  \bibinfo{person}{Fahad~Shahbaz Khan}, {and} \bibinfo{person}{Ming-Hsuan
  Yang}.} \bibinfo{year}{2022}\natexlab{}.
\newblock \showarticletitle{Restormer: Efficient transformer for
  high-resolution image restoration}. In \bibinfo{booktitle}{\emph{Proceedings
  of the IEEE/CVF conference on computer vision and pattern recognition}}.
  \bibinfo{pages}{5728--5739}.
\newblock


\bibitem[Zamir et~al\mbox{.}(2021)]%
        {68}
\bibfield{author}{\bibinfo{person}{Syed~Waqas Zamir}, \bibinfo{person}{Aditya
  Arora}, \bibinfo{person}{Salman Khan}, \bibinfo{person}{Munawar Hayat},
  \bibinfo{person}{Fahad~Shahbaz Khan}, \bibinfo{person}{Ming-Hsuan Yang},
  {and} \bibinfo{person}{Ling Shao}.} \bibinfo{year}{2021}\natexlab{}.
\newblock \showarticletitle{Multi-stage progressive image restoration}. In
  \bibinfo{booktitle}{\emph{Proceedings of the IEEE/CVF conference on computer
  vision and pattern recognition}}. \bibinfo{pages}{14821--14831}.
\newblock


\bibitem[Zeng et~al\mbox{.}(2025)]%
        {39}
\bibfield{author}{\bibinfo{person}{Haijin Zeng}, \bibinfo{person}{Xiangming
  Wang}, \bibinfo{person}{Yongyong Chen}, \bibinfo{person}{Jingyong Su}, {and}
  \bibinfo{person}{Jie Liu}.} \bibinfo{year}{2025}\natexlab{}.
\newblock \showarticletitle{Vision-language gradient descent-driven all-in-one
  deep unfolding networks}. In \bibinfo{booktitle}{\emph{Proceedings of the
  Computer Vision and Pattern Recognition Conference}}.
  \bibinfo{pages}{7524--7533}.
\newblock


\bibitem[Zhang et~al\mbox{.}(2024a)]%
        {3}
\bibfield{author}{\bibinfo{person}{Hua Zhang}, \bibinfo{person}{Liqiang Xiao},
  \bibinfo{person}{Xiaochun Cao}, {and} \bibinfo{person}{Hassan Foroosh}.}
  \bibinfo{year}{2024}\natexlab{a}.
\newblock \showarticletitle{Multiple Adverse Weather Conditions Adaptation for
  Object Detection via Causal Intervention}.
\newblock \bibinfo{journal}{\emph{IEEE Transactions on Pattern Analysis and
  Machine Intelligence}} \bibinfo{volume}{46}, \bibinfo{number}{3}
  (\bibinfo{year}{2024}), \bibinfo{pages}{1742--1756}.
\newblock


\bibitem[Zhang et~al\mbox{.}(2023)]%
        {40}
\bibfield{author}{\bibinfo{person}{Jinghao Zhang}, \bibinfo{person}{Jie Huang},
  \bibinfo{person}{Mingde Yao}, \bibinfo{person}{Zizheng Yang},
  \bibinfo{person}{Hu Yu}, \bibinfo{person}{Man Zhou}, {and}
  \bibinfo{person}{Feng Zhao}.} \bibinfo{year}{2023}\natexlab{}.
\newblock \showarticletitle{Ingredient-oriented multi-degradation learning for
  image restoration}. In \bibinfo{booktitle}{\emph{Proceedings of the IEEE/CVF
  conference on computer vision and pattern recognition}}.
  \bibinfo{pages}{5825--5835}.
\newblock


\bibitem[Zhang et~al\mbox{.}(2017a)]%
        {5}
\bibfield{author}{\bibinfo{person}{Kai Zhang}, \bibinfo{person}{Wangmeng Zuo},
  \bibinfo{person}{Yunjin Chen}, \bibinfo{person}{Deyu Meng}, {and}
  \bibinfo{person}{Lei Zhang}.} \bibinfo{year}{2017}\natexlab{a}.
\newblock \showarticletitle{Beyond a Gaussian Denoiser: Residual Learning of
  Deep CNN for Image Denoising}.
\newblock \bibinfo{journal}{\emph{IEEE Transactions on Image Processing}}
  \bibinfo{volume}{26}, \bibinfo{number}{7} (\bibinfo{year}{2017}),
  \bibinfo{pages}{3142--3155}.
\newblock


\bibitem[Zhang et~al\mbox{.}(2017b)]%
        {su_7}
\bibfield{author}{\bibinfo{person}{Kai Zhang}, \bibinfo{person}{Wangmeng Zuo},
  \bibinfo{person}{Shuhang Gu}, {and} \bibinfo{person}{Lei Zhang}.}
  \bibinfo{year}{2017}\natexlab{b}.
\newblock \showarticletitle{Learning deep CNN denoiser prior for image
  restoration}. In \bibinfo{booktitle}{\emph{Proceedings of the IEEE conference
  on computer vision and pattern recognition}}. \bibinfo{pages}{3929--3938}.
\newblock


\bibitem[Zhang et~al\mbox{.}(2018)]%
        {53}
\bibfield{author}{\bibinfo{person}{Kai Zhang}, \bibinfo{person}{Wangmeng Zuo},
  {and} \bibinfo{person}{Lei Zhang}.} \bibinfo{year}{2018}\natexlab{}.
\newblock \showarticletitle{FFDNet: Toward a Fast and Flexible Solution for
  CNN-Based Image Denoising}.
\newblock \bibinfo{journal}{\emph{IEEE Transactions on Image Processing}}
  \bibinfo{volume}{27}, \bibinfo{number}{9} (\bibinfo{year}{2018}),
  \bibinfo{pages}{4608--4622}.
\newblock


\bibitem[Zhang et~al\mbox{.}(2021)]%
        {55}
\bibfield{author}{\bibinfo{person}{Xiaoqin Zhang}, \bibinfo{person}{Runhua
  Jiang}, \bibinfo{person}{Tao Wang}, {and} \bibinfo{person}{Wenhan Luo}.}
  \bibinfo{year}{2021}\natexlab{}.
\newblock \showarticletitle{Single Image Dehazing via Dual-Path Recurrent
  Network}.
\newblock \bibinfo{journal}{\emph{IEEE Transactions on Image Processing}}
  \bibinfo{volume}{30} (\bibinfo{year}{2021}), \bibinfo{pages}{5211--5222}.
\newblock


\bibitem[Zhang et~al\mbox{.}(2026a)]%
        {41}
\bibfield{author}{\bibinfo{person}{Xu Zhang}, \bibinfo{person}{Jiaqi Ma},
  \bibinfo{person}{Guoli Wang}, \bibinfo{person}{Qian Zhang},
  \bibinfo{person}{Huan Zhang}, {and} \bibinfo{person}{Lefei Zhang}.}
  \bibinfo{year}{2026}\natexlab{a}.
\newblock \showarticletitle{Perceive-IR: Learning to Perceive Degradation
  Better for All-in-One Image Restoration}.
\newblock \bibinfo{journal}{\emph{IEEE Transactions on Image Processing}}
  \bibinfo{volume}{35} (\bibinfo{year}{2026}), \bibinfo{pages}{2018--2033}.
\newblock


\bibitem[Zhang et~al\mbox{.}(2026b)]%
        {78}
\bibfield{author}{\bibinfo{person}{Xu Zhang}, \bibinfo{person}{Huan Zhang},
  \bibinfo{person}{Guoli Wang}, \bibinfo{person}{Qian Zhang}, {and}
  \bibinfo{person}{Lefei Zhang}.} \bibinfo{year}{2026}\natexlab{b}.
\newblock \showarticletitle{ClearAIR: A Human-Visual-Perception-Inspired
  All-in-One Image Restoration}.
\newblock \bibinfo{journal}{\emph{Proceedings of the AAAI Conference on
  Artificial Intelligence}} \bibinfo{volume}{40}, \bibinfo{number}{15}
  (\bibinfo{year}{2026}), \bibinfo{pages}{12861--12869}.
\newblock


\bibitem[Zhang et~al\mbox{.}(2025)]%
        {80}
\bibfield{author}{\bibinfo{person}{Yuhong Zhang}, \bibinfo{person}{Hengsheng
  Zhang}, \bibinfo{person}{Xinning Chai}, \bibinfo{person}{Zhengxue Cheng},
  \bibinfo{person}{Rong Xie}, \bibinfo{person}{Li Song}, {and}
  \bibinfo{person}{Wenjun Zhang}.} \bibinfo{year}{2025}\natexlab{}.
\newblock \showarticletitle{Diff-Restorer: Unleashing Visual Prompts for
  Diffusion-based Universal Image Restoration}.
\newblock \bibinfo{journal}{\emph{IEEE Transactions on Circuits and Systems for
  Video Technology}} (\bibinfo{year}{2025}), \bibinfo{pages}{1--1}.
\newblock


\bibitem[Zhang et~al\mbox{.}(2024b)]%
        {20}
\bibfield{author}{\bibinfo{person}{Yafei Zhang}, \bibinfo{person}{Shen Zhou},
  {and} \bibinfo{person}{Huafeng Li}.} \bibinfo{year}{2024}\natexlab{b}.
\newblock \showarticletitle{Depth information assisted collaborative mutual
  promotion network for single image dehazing}. In
  \bibinfo{booktitle}{\emph{Proceedings of the IEEE/CVF conference on computer
  vision and pattern recognition}}. \bibinfo{pages}{2846--2855}.
\newblock


\bibitem[Zhu et~al\mbox{.}(2023)]%
        {71}
\bibfield{author}{\bibinfo{person}{Yurui Zhu}, \bibinfo{person}{Tianyu Wang},
  \bibinfo{person}{Xueyang Fu}, \bibinfo{person}{Xuanyu Yang},
  \bibinfo{person}{Xin Guo}, \bibinfo{person}{Jifeng Dai}, \bibinfo{person}{Yu
  Qiao}, {and} \bibinfo{person}{Xiaowei Hu}.} \bibinfo{year}{2023}\natexlab{}.
\newblock \showarticletitle{Learning Weather-General and Weather-Specific
  Features for Image Restoration Under Multiple Adverse Weather Conditions}. In
  \bibinfo{booktitle}{\emph{2023 IEEE/CVF Conference on Computer Vision and
  Pattern Recognition (CVPR)}}. \bibinfo{pages}{21747--21758}.
\newblock


\bibitem[Özdenizci and Legenstein(2023)]%
        {72}
\bibfield{author}{\bibinfo{person}{Ozan Özdenizci} {and}
  \bibinfo{person}{Robert Legenstein}.} \bibinfo{year}{2023}\natexlab{}.
\newblock \showarticletitle{Restoring Vision in Adverse Weather Conditions With
  Patch-Based Denoising Diffusion Models}.
\newblock \bibinfo{journal}{\emph{IEEE Transactions on Pattern Analysis and
  Machine Intelligence}} \bibinfo{volume}{45}, \bibinfo{number}{8}
  (\bibinfo{year}{2023}), \bibinfo{pages}{10346--10357}.
\newblock


\end{thebibliography}

\clearpage
\appendix
\twocolumn[
\begin{@twocolumnfalse}
	\centering
	{\LARGE\bfseries Supplementary Material\\
		QuReC: All-in-One Image Restoration with Query-Specific Guidance and Local-Global Response Calibration\par}
	\vspace{1em}
\end{@twocolumnfalse}
]

\noindent\textbf{This supplementary material is organized as follows:}\\
\textbf{Section~A} presents the training objective of QuReC.\\
\textbf{Section~B} provides sensitivity analyses on several key design choices.\\
\textbf{Section~C} reports additional experimental results, including perceptual quality comparisons under the all-in-one and composite-degradation settings, quantitative comparisons under dedicated single-task settings, and more qualitative visual comparisons.\\
\textbf{Section~D} provides additional analyses and methodological details, including computational overhead, robustness to unseen degradations, weakly supervised prototype matching, and spatial visualizations of query-wise prototype responses.

\section{Training Objective}

The overall training objective of QuReC consists of a restoration loss and a prototype matching learning objective. Specifically, the restoration branch is optimized with a pixel-wise reconstruction loss together with a frequency-domain loss:
\begin{equation}
	\mathcal{L}_{\mathrm{res}}
	=
	\mathcal{L}_{1}
	+
	0.1\,\mathcal{L}_{\mathrm{fft}},
\end{equation}
where $\mathcal{L}_{1}$ denotes the $\ell_1$ reconstruction loss and $\mathcal{L}_{\mathrm{fft}}$ denotes the frequency-domain supervision term.

To stabilize query-wise prototype matching, we further introduce the prototype matching learning objective:
\begin{equation}
	\mathcal{L}_{\mathrm{pm}}
	=
	\lambda_{\mathrm{bal}}\mathcal{L}_{\mathrm{bal}}
	+
	\lambda_{\mathrm{match}}\mathcal{L}_{\mathrm{match}},
\end{equation}
where $\mathcal{L}_{\mathrm{bal}}$ is the mixed load-balancing loss and $\mathcal{L}_{\mathrm{match}}$ is the weak matching supervision loss. The overall training objective is then formulated as:
\begin{equation}
	\mathcal{L}
	=
	\mathcal{L}_{\mathrm{res}}
	+
	\mathcal{L}_{\mathrm{pm}}.
\end{equation}

Among them, the mixed load-balancing term jointly considers soft prototype usage and hard prototype occupancy:
\begin{equation}
	\mathcal{L}_{\mathrm{bal}}
	=
	\alpha_s \mathcal{L}_{\mathrm{soft}}
	+
	\alpha_h \mathcal{L}_{\mathrm{hard}},
\end{equation}
where $\alpha_s$ and $\alpha_h$ denote the coefficients for the soft and hard balancing terms, respectively. In all experiments, the default setting uses $\lambda_{\mathrm{bal}}=0.01$, $\lambda_{\mathrm{match}}=0.05$, $\alpha_s=0.7$, and $\alpha_h=0.3$. In the following, we further analyze the sensitivity of QuReC to these design choices.

\section{Sensitivity Analysis}
In this section, we provide additional sensitivity studies to further examine the robustness of QuReC with respect to several key design choices, including prototype prompt formulation, local window size, and the coefficients in the prototype matching learning objective. Unless otherwise specified, all sensitivity studies in this section are conducted under the same three-task all-in-one evaluation protocol, and the default configuration corresponds to the full QuReC model reported in the main paper.

\begin{table}[t]
	\centering
	\small
	\setlength{\tabcolsep}{8pt}
	\renewcommand{\arraystretch}{1.0}
	\caption{Sensitivity to prototype prompt formulation under the three-task all-in-one setting.}
	\label{tab_sup_1}
	\begin{tabular}{lcc}
		\toprule
		\textbf{Prompt formulation} & \textbf{PSNR} $\uparrow$ & \textbf{SSIM} $\uparrow$ \\
		\midrule
		\cellcolor[HTML]{ffc7ce}{Single short prompt} & \cellcolor[HTML]{ffc7ce}{33.16} & \cellcolor[HTML]{ffc7ce}{0.920} \\
		\cellcolor[HTML]{DFE8FE}{Longer descriptive prompt} & \cellcolor[HTML]{DFE8FE}{33.14} & \cellcolor[HTML]{DFE8FE}{0.920} \\
		Multi-prompt averaging & 33.11 & 0.919 \\
		\bottomrule
	\end{tabular}
\end{table}

\subsection{Sensitivity to Prototype Prompt Formulation}
Since DQRM relies on a CLIP-based prototype bank to provide degradation-aware query guidance, the wording of the text prototypes may affect the quality of the resulting semantic anchors. To study this issue, we compare three prototype construction strategies under the three-task all-in-one setting:
(1) \emph{single short prompt}, which uses compact category names, namely ``Noise'', ``Rain'', and ``Haze'';
(2) \emph{longer descriptive prompt}, which uses more explicit sentence templates, namely ``a degraded image with noise'', ``a degraded image with rain streaks'', and ``a degraded image with haze''; and
(3) \emph{multi-prompt averaging}, which averages the CLIP embeddings of multiple synonymous prompts for each category, namely \{``Noise'', ``a noisy image'', ``a degraded image with noise''\}, \{``Rain'', ``a rainy image'', ``a degraded image with rain streaks''\}, and \{``Haze'', ``a hazy image'', ``a degraded image with haze''\}. Note that these prompt formulations are only used for the three-task setting considered in this sensitivity study, while other task settings use their own setting-dependent prototype vocabularies. The quantitative results are reported in Table~\ref{tab_sup_1}. It can be observed that the performance differences among the three prompt formulations are marginal. In particular, the default short-prompt setting achieves the best result, while the longer descriptive prompt and the multi-prompt averaging strategy lead to highly comparable performance. This suggests that QuReC is not sensitive to elaborate prompt engineering, and that compact category-level prompts are already sufficient to provide stable semantic anchors for degradation-aware query reconstruction.

\subsection{Sensitivity to Local Window Size}
We further analyze the sensitivity of QuReC to the local window size in LGRCM. The local window size controls the spatial range of neighborhood aggregation in the local branch and therefore affects the balance between local specificity and contextual coverage. To this end, we evaluate several window sizes while keeping other settings unchanged. Table~\ref{tab_sup_2} reports the corresponding results. We can see that QuReC remains stable across different local window sizes, while the default setting with a relatively small local window achieves the best performance. A possible reason is that smaller local windows are more suitable for capturing fine-grained neighborhood structures under spatially heterogeneous degradations, whereas overly large local windows may weaken locality and blur the distinction between nearby regions with different degradation patterns. Overall, the limited variation across different settings indicates that QuReC is reasonably robust to the choice of local window size.
\begin{table}[t]
	\centering
	\small
	\setlength{\tabcolsep}{12pt}
	\renewcommand{\arraystretch}{0.9}
	\caption{Sensitivity to the local window size in LGRCM under the three-task all-in-one setting.}
	\label{tab_sup_2}
	\begin{tabular}{c|c|c}
		\toprule
		\textbf{Window size} $k$ & \textbf{PSNR} $\uparrow$ & \textbf{SSIM} $\uparrow$ \\
		\midrule
		\cellcolor[HTML]{ffc7ce}{3} & \cellcolor[HTML]{ffc7ce}{33.16} & \cellcolor[HTML]{ffc7ce}{0.920} \\
		\cellcolor[HTML]{DFE8FE}{5} & \cellcolor[HTML]{DFE8FE}{33.13} & \cellcolor[HTML]{DFE8FE}{0.919} \\
		7 & 33.08 & 0.919 \\
		\bottomrule
	\end{tabular}
\end{table}
\begin{table}[h]
	\centering
	\small
	\setlength{\tabcolsep}{12pt}
	\renewcommand{\arraystretch}{1.0}
	\caption{Sensitivity to $\lambda_{\mathrm{bal}}$ and $\lambda_{\mathrm{match}}$ under the three-task all-in-one setting.}
	\label{tab_sup_3}
	\begin{tabular}{c|c|c|c}
		\toprule
		$\lambda_{\mathrm{bal}}$ & $\lambda_{\mathrm{match}}$ & \textbf{PSNR} $\uparrow$ & \textbf{SSIM} $\uparrow$ \\
		\midrule
		0.01 & 0.03 & 33.12 & 0.919 \\
		\cellcolor[HTML]{ffc7ce}{0.01} & \cellcolor[HTML]{ffc7ce}{0.05} & \cellcolor[HTML]{ffc7ce}{33.16} & \cellcolor[HTML]{ffc7ce}{0.920} \\
		0.01 & 0.08 & 33.12 & 0.919 \\
		\midrule
		0.005 & 0.05 & 33.11 & 0.919 \\
		\cellcolor[HTML]{DFE8FE}{0.02} & \cellcolor[HTML]{DFE8FE}{0.05} & \cellcolor[HTML]{DFE8FE}{33.14} & \cellcolor[HTML]{DFE8FE}{0.920} \\
		0.005 & 0.03 & 33.09 & 0.918 \\
		\bottomrule
	\end{tabular}
\end{table}

\subsection{Sensitivity to $\lambda_{\mathrm{bal}}$ and $\lambda_{\mathrm{match}}$}
We next investigate the sensitivity of QuReC to the coefficients of the prototype matching learning objective. Specifically, $\lambda_{\mathrm{bal}}$ controls the strength of the prototype usage balancing regularization, while $\lambda_{\mathrm{match}}$ controls the strength of the weak semantic matching supervision. Both coefficients influence the stability of query-wise prototype assignment during training. The results are summarized in Table~\ref{tab_sup_3}. As shown in the table, QuReC achieves the best performance around the default setting and remains relatively stable within a reasonable coefficient range. When the regularization weights are set to smaller values, the prototype matching process becomes less constrained, which weakens the benefit of degradation-aware routing. In contrast, when the coefficients are increased, the auxiliary objective does not bring further gains and may slightly affect the restoration objective. These results indicate that the default setting provides a good balance between restoration optimization and prototype matching regularization.

\begin{table}[t]
	\centering
	\small
	\setlength{\tabcolsep}{12pt}
	\renewcommand{\arraystretch}{1.0}
	\caption{Sensitivity to $\alpha_s$ and $\alpha_h$ under the three-task all-in-one setting.}
	\label{tab_sup_4}
	\begin{tabular}{cccc}
		\toprule
		$\alpha_s$ & $\alpha_h$ & \textbf{PSNR} $\uparrow$ & \textbf{SSIM} $\uparrow$ \\
		\midrule
		1.0 & 0.0 & 33.10 & 0.919 \\
		\cellcolor[HTML]{ffc7ce}{0.7} & \cellcolor[HTML]{ffc7ce}{0.3} & \cellcolor[HTML]{ffc7ce}{33.16} & \cellcolor[HTML]{ffc7ce}{0.920} \\
		\cellcolor[HTML]{DFE8FE}{0.5} & \cellcolor[HTML]{DFE8FE}{0.5} & \cellcolor[HTML]{DFE8FE}{33.15} & \cellcolor[HTML]{DFE8FE}{0.920} \\
		0.3 & 0.7 & 33.12 & 0.919 \\
		0.0 & 1.0 & 33.09 & 0.919 \\
		\bottomrule
	\end{tabular}
\end{table}

\subsection{Sensitivity to $\alpha_s$ and $\alpha_h$}
Finally, we study the sensitivity of QuReC to the weighting coefficients of the soft and hard balancing terms in $\mathcal{L}_{\mathrm{bal}}$. Here, $\alpha_s$ emphasizes balancing at the probability level, while $\alpha_h$ emphasizes balancing at the discrete assignment level. A proper combination of the two is expected to improve prototype utilization without introducing unnecessary instability. As reported in Table~\ref{tab_sup_4}, QuReC is not highly sensitive to moderate changes in $\alpha_s$ and $\alpha_h$, and the default setting $(\alpha_s,\alpha_h)=(0.7,0.3)$ achieves the best performance. Using only one type of balancing signal is less effective than combining both of them, which indicates that the soft and hard balancing terms play complementary roles. A possible reason is that the soft term promotes smoother global prototype usage, whereas the hard term further constrains discrete token-level occupancy. Their combination therefore leads to more balanced and stable prototype matching, which in turn benefits unified restoration performance.

\section{Additional Experimental Results}
In this supplementary material, we provide additional quantitative and qualitative results to further validate the effectiveness of the proposed QuReC under all-in-one, composite-degradation, and dedicated single-task restoration settings.
\subsection{Perceptual Quality Comparison under the All-in-One Setting with Three Degradations}
Table~\ref{tab_sup_5} reports the perceptual quality comparison under the all-in-one setting with three degradation tasks, including denoising, deraining, and dehazing. We adopt LPIPS and DISTS as perceptual quality metrics, where lower values indicate better perceptual fidelity. As shown in Table~\ref{tab_sup_5}, QuReC consistently achieves the best performance across all five evaluation settings and obtains the lowest average LPIPS and DISTS among all compared methods. Specifically, QuReC achieves LPIPS/DISTS scores of 0.0528/0.0708, 0.0914/0.0998, and 0.1797/0.1486 on BSD68 with noise levels 15, 25, and 50, respectively. It also achieves the best perceptual results on Rain100L and SOTS, with LPIPS/DISTS of 0.0130/0.0158 and 0.2086/0.0400, respectively. These results indicate that QuReC not only improves restoration fidelity in a distortion-oriented sense, but also produces perceptually more natural and visually pleasing outputs. This advantage can be attributed to the proposed query-specific degradation guidance and local-global response calibration, which enable the model to more effectively suppress degradations while preserving visually important structures and textures.
\begin{table*}[h]
	\centering
	\small
	\setlength{\tabcolsep}{5.5pt}
	\renewcommand{\arraystretch}{1.1}
	\caption{Performance Comparison of Different Methods on Three Degradation Tasks. The best and second-best performances for each metric are highlighted with \colorbox[HTML]{ffc7ce}{Red} and \colorbox[HTML]{DFE8FE}{Blue} backgrounds.}
	\label{tab_sup_5}  
	\begin{tabular}{c|cc|cc|cc|cc|cc|cc} 
		\toprule
		
		\multicolumn{1}{c|}{\multirow{3}{*}[-1.5ex]{\textbf{Method}}}
		& \multicolumn{6}{c|}{\textbf{Denoising}}
		& \multicolumn{2}{c|}{\textbf{Deraining}}
		& \multicolumn{2}{c|}{\textbf{Dehazing}}
		& \multicolumn{2}{c}{\multirow{2}{*}[-0.5ex]{\textbf{Average}}} \\
		
		\cmidrule{2-11}
		
		& \multicolumn{2}{c|}{$\mathrm{BSD68}_{\sigma=15}$}
		& \multicolumn{2}{c|}{$\mathrm{BSD68}_{\sigma=25}$}
		& \multicolumn{2}{c|}{$\mathrm{BSD68}_{\sigma=50}$}
		& \multicolumn{2}{c|}{Rain100L}
		& \multicolumn{2}{c|}{SOTS}
		& \multicolumn{2}{c}{} \\
		
		\cmidrule{2-13}
		
		& $\mathrm{LPIPS}\downarrow$ & $\mathrm{DISTS}\downarrow$
		& $\mathrm{LPIPS}\downarrow$ & $\mathrm{DISTS}\downarrow$
		& $\mathrm{LPIPS}\downarrow$ & $\mathrm{DISTS}\downarrow$
		& $\mathrm{LPIPS}\downarrow$ & $\mathrm{DISTS}\downarrow$
		& $\mathrm{LPIPS}\downarrow$ & $\mathrm{DISTS}\downarrow$
		& $\mathrm{LPIPS}\downarrow$ & $\mathrm{DISTS}\downarrow$ \\
		\midrule
		
		AirNet~\cite{33} 
		& 0.0648 & 0.0884
		& 0.1134 & 0.1230
		& 0.2083 & 0.1721
		& 0.0306 & 0.0386
		& 0.2273 & 0.0624
		& 0.1289 & 0.0969 \\
		
		PromptIR~\cite{36} 
		& 0.0662 & 0.0883
		& 0.1148 & 0.1215
		& 0.2151 & 0.1723
		& 0.0191 & 0.0268
		& 0.2160 & 0.0451
		& 0.1262 & 0.0908 \\
		
		AdaIR~\cite{59} 
		& 0.0634 & 0.0853
		& 0.1098 & 0.1198
		& 0.2128 & 0.1704
		& \cellcolor[HTML]{DFE8FE}{0.0147} & \cellcolor[HTML]{DFE8FE}{0.0187}
		& 0.2171 & 0.0423
		& 0.1236 & 0.0873 \\
		
		MoCE-IR~\cite{44} 
		& \cellcolor[HTML]{DFE8FE}{0.0598} & \cellcolor[HTML]{DFE8FE}{0.0754}
		& 0.1029 & 0.1054
		& \cellcolor[HTML]{DFE8FE}{0.1927} & \cellcolor[HTML]{DFE8FE}{0.1500}
		& \cellcolor[HTML]{DFE8FE}{0.0147} & 0.0190
		& \cellcolor[HTML]{DFE8FE}{0.2120} & \cellcolor[HTML]{DFE8FE}{0.0417}
		& \cellcolor[HTML]{DFE8FE}{0.1164} & \cellcolor[HTML]{DFE8FE}{0.0783} \\
		
		DFPIR~\cite{42} 
		& 0.0633 & 0.0817
		& \cellcolor[HTML]{DFE8FE}{0.0927} & \cellcolor[HTML]{DFE8FE}{0.1003}
		& 0.1965 & 0.1575
		& 0.0153 & 0.0209
		& 0.2165 & 0.0430
		& 0.1169 & 0.0807 \\
		
		\midrule
		
		QuReC (Ours) 
		& \cellcolor[HTML]{ffc7ce}{0.0528} & \cellcolor[HTML]{ffc7ce}{0.0708}
		& \cellcolor[HTML]{ffc7ce}{0.0914} & \cellcolor[HTML]{ffc7ce}{0.0998}
		& \cellcolor[HTML]{ffc7ce}{0.1797} & \cellcolor[HTML]{ffc7ce}{0.1486}
		& \cellcolor[HTML]{ffc7ce}{0.0130} & \cellcolor[HTML]{ffc7ce}{0.0158}
		& \cellcolor[HTML]{ffc7ce}{0.2086} & \cellcolor[HTML]{ffc7ce}{0.0400}
		& \cellcolor[HTML]{ffc7ce}{0.1091} & \cellcolor[HTML]{ffc7ce}{0.0750} \\
		
		\bottomrule
	\end{tabular}
\end{table*}

\begin{table*}[h]
	\centering
	\small
	\setlength{\tabcolsep}{3pt}
	\renewcommand{\arraystretch}{1.1}
	\caption{Performance Comparison of Different Methods on the CDD11 dataset in terms of LPIPS. The best and second-best performances for each metric are highlighted with \colorbox[HTML]{ffc7ce}{Red} and \colorbox[HTML]{DFE8FE}{Blue} backgrounds.}
	\label{tab_sup_6}  
	\begin{tabularx}{\linewidth}{c|*{11}{>{\centering\arraybackslash}X|}>{\centering\arraybackslash}X} 
		\toprule
		
		\multicolumn{1}{c|}{\multirow{2}{*}[-0.8ex]{\textbf{Method}}}
		& \textbf{Low (L)}
		& \textbf{Haze (H)}
		& \textbf{Rain (R)}
		& \textbf{Snow (S)}
		& \textbf{L+H}
		& \textbf{L+R}
		& \textbf{L+S}
		& \textbf{H+R}
		& \textbf{H+S}
		& \textbf{L+H+R}
		& \textbf{L+H+S}
		& \textbf{Average} \\
		
		\cmidrule{2-13}
		
		& $\mathrm{LPIPS}\downarrow$ & $\mathrm{LPIPS}\downarrow$ & $\mathrm{LPIPS}\downarrow$ & $\mathrm{LPIPS}\downarrow$ & $\mathrm{LPIPS}\downarrow$ & $\mathrm{LPIPS}\downarrow$ & $\mathrm{LPIPS}\downarrow$ & $\mathrm{LPIPS}\downarrow$ & $\mathrm{LPIPS}\downarrow$ & $\mathrm{LPIPS}\downarrow$ & $\mathrm{LPIPS}\downarrow$ & $\mathrm{LPIPS}\downarrow$ \\
		\midrule
		
		AirNet~\cite{33} 
		& 0.4088 & 0.1223 & 0.2674 & 0.2502 & 0.5228 & 0.5794 & 0.5775 & 0.3568 & 0.2739 & 0.6250 & 0.6080 & 0.4175 \\
		
		PromptIR~\cite{36} 
		& 0.3882 & 0.0486 & 0.1731 & 0.1987 & 0.5074 & 0.5635 & 0.5703 & 0.3209 & 0.2293 & 0.6466 & 0.6250 & 0.3883 \\
		
		OneRestore~\cite{67} 
		& \cellcolor[HTML]{DFE8FE}{0.1616} & \cellcolor[HTML]{DFE8FE}{0.0072} & 0.0348 & 0.0225 & \cellcolor[HTML]{DFE8FE}{0.1710} & \cellcolor[HTML]{DFE8FE}{0.1951} & \cellcolor[HTML]{DFE8FE}{0.1962} & \cellcolor[HTML]{DFE8FE}{0.0382} & \cellcolor[HTML]{DFE8FE}{0.0306} & \cellcolor[HTML]{DFE8FE}{0.1917} & \cellcolor[HTML]{DFE8FE}{0.1947} & \cellcolor[HTML]{DFE8FE}{0.1131} \\
		
		MoCE-IR~\cite{44} 
		& 0.1692 & 0.0095 & \cellcolor[HTML]{DFE8FE}{0.0335} & \cellcolor[HTML]{DFE8FE}{0.0219} & 0.1798 & 0.1979 & 0.1998 & 0.0362 & 0.0342 & 0.2131 & 0.2023 & 0.1179 \\
		
		\midrule
		
		QuReC (Ours) 
		& \cellcolor[HTML]{ffc7ce}{0.1555} & \cellcolor[HTML]{ffc7ce}{0.0046} & \cellcolor[HTML]{ffc7ce}{0.0258} & \cellcolor[HTML]{ffc7ce}{0.0154} & \cellcolor[HTML]{ffc7ce}{0.1625} & \cellcolor[HTML]{ffc7ce}{0.1746} & \cellcolor[HTML]{ffc7ce}{0.1721} & \cellcolor[HTML]{ffc7ce}{0.0286} & \cellcolor[HTML]{ffc7ce}{0.0204} & \cellcolor[HTML]{ffc7ce}{0.1905} & \cellcolor[HTML]{ffc7ce}{0.1780} & \cellcolor[HTML]{ffc7ce}{0.1026} \\
		
		\bottomrule
	\end{tabularx}
\end{table*}

\begin{table*}[h]
	\centering
	\small
	\setlength{\tabcolsep}{3pt}
	\renewcommand{\arraystretch}{1}
	\caption{Performance Comparison of Different Methods on the CDD11 dataset in terms of DISTS. The best and second-best performances for each metric are highlighted with \colorbox[HTML]{ffc7ce}{Red} and \colorbox[HTML]{DFE8FE}{Blue} backgrounds.}
	\label{tab_sup_7}  
	\begin{tabularx}{\linewidth}{c|*{11}{>{\centering\arraybackslash}X|}>{\centering\arraybackslash}X} 
		\toprule
		
		\multicolumn{1}{c|}{\multirow{2}{*}[-0.8ex]{\textbf{Method}}}
		& \textbf{Low (L)}
		& \textbf{Haze (H)}
		& \textbf{Rain (R)}
		& \textbf{Snow (S)}
		& \textbf{L+H}
		& \textbf{L+R}
		& \textbf{L+S}
		& \textbf{H+R}
		& \textbf{H+S}
		& \textbf{L+H+R}
		& \textbf{L+H+S}
		& \textbf{Average} \\
		
		\cmidrule{2-13}
		
		& $\mathrm{DISTS}\downarrow$ & $\mathrm{DISTS}\downarrow$ & $\mathrm{DISTS}\downarrow$ & $\mathrm{DISTS}\downarrow$ & $\mathrm{DISTS}\downarrow$ & $\mathrm{DISTS}\downarrow$ & $\mathrm{DISTS}\downarrow$ & $\mathrm{DISTS}\downarrow$ & $\mathrm{DISTS}\downarrow$ & $\mathrm{DISTS}\downarrow$ & $\mathrm{DISTS}\downarrow$ & $\mathrm{DISTS}\downarrow$ \\
		\midrule
		
		AirNet~\cite{33} 
		& 0.2467 & 0.0693 & 0.1603 & 0.1381 & 0.2684 & 0.3059 & 0.2951 & 0.2109 & 0.1523 & 0.3370 & 0.3021 & 0.2260 \\
		
		PromptIR~\cite{36} 
		& 0.2407 & 0.0420 & 0.1214 & 0.1249 & 0.3041 & 0.3076 & 0.2904 & 0.2008 & 0.1365 & 0.3669 & 0.3311 & 0.2242 \\
		
		OneRestore~\cite{67} 
		& \cellcolor[HTML]{DFE8FE}{0.1191} & \cellcolor[HTML]{DFE8FE}{0.0081} & \cellcolor[HTML]{DFE8FE}{0.0307} & \cellcolor[HTML]{DFE8FE}{0.0200} & \cellcolor[HTML]{DFE8FE}{0.1225} & 0.1378 & \cellcolor[HTML]{ffc7ce}{0.1173} & \cellcolor[HTML]{DFE8FE}{0.0330} & \cellcolor[HTML]{DFE8FE}{0.0257} & \cellcolor[HTML]{DFE8FE}{0.1371} & \cellcolor[HTML]{DFE8FE}{0.1178} & \cellcolor[HTML]{DFE8FE}{0.0790} \\
		
		MoCE-IR~\cite{44} 
		& 0.1217 & 0.0096 & 0.0361 & 0.0205 & 0.1268 & \cellcolor[HTML]{DFE8FE}{0.1348} & \cellcolor[HTML]{DFE8FE}{0.1331} & 0.0376 & 0.0299 & 0.1448 & 0.1357 & 0.0846 \\
		
		\midrule
		
		QuReC (Ours) 
		& \cellcolor[HTML]{ffc7ce}{0.1149} & \cellcolor[HTML]{ffc7ce}{0.0060} & \cellcolor[HTML]{ffc7ce}{0.0258} & \cellcolor[HTML]{ffc7ce}{0.0152} & \cellcolor[HTML]{ffc7ce}{0.1163} & \cellcolor[HTML]{ffc7ce}{0.1204} & \cellcolor[HTML]{ffc7ce}{0.1173} & \cellcolor[HTML]{ffc7ce}{0.0281} & \cellcolor[HTML]{ffc7ce}{0.0197} & \cellcolor[HTML]{ffc7ce}{0.1293} & \cellcolor[HTML]{ffc7ce}{0.1219} & \cellcolor[HTML]{ffc7ce}{0.0741} \\
		
		\bottomrule
	\end{tabularx}
\end{table*}
\subsection{Perceptual Quality Comparison under the Composite Degradation Setting}
Table~\ref{tab_sup_6} and Table~\ref{tab_sup_7} further report the perceptual quality comparison on the CDD11 benchmark under the composite degradation setting. Table~\ref{tab_sup_6} presents LPIPS results, while Table~\ref{tab_sup_7} presents DISTS results. As shown in both tables, QuReC achieves the best overall performance across the 11 degradation subsets and obtains the lowest average scores among all compared methods. In terms of LPIPS, QuReC achieves the best average result of 0.1026, outperforming the previous best-performing method OneRestore by 0.0105. Moreover, its superiority is consistently observed not only on single degradations such as Low, Haze, Rain, and Snow, but also on more challenging mixed degradations, including L+H, L+R, L+S, H+R, H+S, L+H+R, and L+H+S. In terms of DISTS, QuReC again achieves the best average score of 0.0741 and obtains the lowest values on almost all subsets, while achieving tied-best performance on L+S. These results demonstrate that QuReC is particularly effective in handling spatially complex and compositionally coupled degradations, where reliable query-wise degradation modeling and robust local-global aggregation are especially important.

Figure~\ref{figure5} further provides qualitative comparisons across all 11 degradation cases in the CDD11 dataset. As can be seen, QuReC consistently achieves more visually pleasing restoration results under both single and composite degradation settings. In particular, it removes complex mixed degradations more thoroughly while better preserving structural details and natural image appearance. Compared with previous methods, QuReC produces fewer residual artifacts and less detail distortion in challenging cases such as coupled low-light, haze, rain, and snow degradations. These visual results are consistent with the perceptual metric comparisons in Tables~\ref{tab_sup_6} and \ref{tab_sup_7}, further demonstrating the effectiveness of QuReC under diverse composite degradation scenarios.
\begin{table*}[h]
	\centering
	\small
	\setlength{\tabcolsep}{5pt}
	\renewcommand{\arraystretch}{1}
	\caption{Quantitative comparison of different dehazing methods in terms of PSNR and SSIM. The best and second-best performances for each metric are highlighted with \colorbox[HTML]{ffc7ce}{Red} and \colorbox[HTML]{DFE8FE}{Blue} backgrounds.}
	\label{tab_sup_8}
	\begin{tabular}{c|ccccccccc}
		\toprule
		\textbf{Method} & DehazeNet~\cite{46} & MSCNN~\cite{su_1} & EPDN~\cite{su_2} & FDGAN~\cite{su_3} & AirNet~\cite{33} & Restormer~\cite{31} & PromptIR~\cite{36} & DFPIR~\cite{42} & QuReC (Ours) \\
		\midrule
		PSNR & 22.46 & 22.06 & 22.57 & 23.15 & 23.18 & 30.87 & 31.31 & \cellcolor[HTML]{DFE8FE}{32.00} & \cellcolor[HTML]{ffc7ce}{32.13} \\
		SSIM & 0.851 & 0.908 & 0.863 & 0.921 & 0.900 & 0.969 & 0.973 & \cellcolor[HTML]{DFE8FE}{0.981} & \cellcolor[HTML]{ffc7ce}{0.983} \\
		\bottomrule
	\end{tabular}
\end{table*}

\begin{table*}[h]
	\centering
	\small
	\setlength{\tabcolsep}{6.5pt}
	\renewcommand{\arraystretch}{1}
	\caption{Quantitative comparison of different deraining methods in terms of PSNR and SSIM. The best and second-best performances for each metric are highlighted with \colorbox[HTML]{ffc7ce}{Red} and \colorbox[HTML]{DFE8FE}{Blue} backgrounds.}
	\label{tab_sup_9}
	\begin{tabular}{c|ccccccccc}
		\toprule
		\textbf{Method} & UMR~\cite{su_4} & SIRR~\cite{su_5} & MSPFN~\cite{15} & LPNet~\cite{su_6} & AirNet~\cite{33} & Restormer~\cite{31} & PromptIR~\cite{36} & DFPIR~\cite{42} & QuReC (Ours) \\
		\midrule
		PSNR & 32.39 & 32.37 & 33.50 & 33.61 & 34.90 & 36.74 & 37.04 & \cellcolor[HTML]{DFE8FE}{39.08} & \cellcolor[HTML]{ffc7ce}{39.14} \\
		SSIM & 0.921 & 0.926 & 0.948 & 0.958 & 0.977 & 0.978 & 0.979 & \cellcolor[HTML]{DFE8FE}{0.984} & \cellcolor[HTML]{ffc7ce}{0.986} \\
		\bottomrule
	\end{tabular}
\end{table*}

\begin{table*}[h]
	\centering
	\small
	\setlength{\tabcolsep}{6.5pt}
	\renewcommand{\arraystretch}{1}
	\caption{Performance comparison of different methods on image denoising benchmarks. The best and second-best performances for each metric are highlighted with \colorbox[HTML]{ffc7ce}{Red} and \colorbox[HTML]{DFE8FE}{Blue} backgrounds.}
	\label{tab_sup_10}
	\begin{tabular}{c|cc|cc|cc|cc|cc|cc}
		\toprule
		
		\multicolumn{1}{c|}{\multirow{3}{*}[-1.5ex]{\textbf{Method}}}
		& \multicolumn{6}{c|}{\textbf{Denoising on CBSD68}}
		& \multicolumn{6}{c}{\textbf{Denoising on Urban100}} \\
		
		\cmidrule{2-13}
		
		& \multicolumn{2}{c|}{$\sigma=15$}
		& \multicolumn{2}{c|}{$\sigma=25$}
		& \multicolumn{2}{c|}{$\sigma=50$}
		& \multicolumn{2}{c|}{$\sigma=15$}
		& \multicolumn{2}{c|}{$\sigma=25$}
		& \multicolumn{2}{c}{$\sigma=50$} \\
		
		\cmidrule{2-13}
		
		& $\mathrm{PSNR}\uparrow$ & $\mathrm{SSIM}\uparrow$
		& $\mathrm{PSNR}\uparrow$ & $\mathrm{SSIM}\uparrow$
		& $\mathrm{PSNR}\uparrow$ & $\mathrm{SSIM}\uparrow$
		& $\mathrm{PSNR}\uparrow$ & $\mathrm{SSIM}\uparrow$
		& $\mathrm{PSNR}\uparrow$ & $\mathrm{SSIM}\uparrow$
		& $\mathrm{PSNR}\uparrow$ & $\mathrm{SSIM}\uparrow$ \\
		\midrule
		
		IRCNN~\cite{su_7}
		& 33.87 & 0.929
		& 31.18 & 0.882
		& 27.88 & 0.790
		& 27.59 & 0.833
		& 31.20 & 0.909
		& 27.70 & 0.840 \\
		
		FFDNet~\cite{53}
		& 33.87 & 0.929
		& 31.21 & 0.882
		& 27.96 & 0.789
		& 33.83 & 0.942
		& 31.40 & 0.912
		& 28.05 & 0.848 \\
		
		BRDNet~\cite{su_8}
		& 34.10 & 0.929
		& 31.43 & 0.885
		& 28.16 & 0.794
		& 34.42 & 0.946
		& 31.99 & 0.919
		& 28.56 & 0.858 \\
		
		AirNet~\cite{33}
		& 34.14 & \cellcolor[HTML]{DFE8FE}{0.936}
		& 31.48 & 0.893
		& 28.23 & 0.806
		& 34.40 & \cellcolor[HTML]{DFE8FE}{0.949}
		& 32.10 & 0.924
		& 28.88 & 0.871 \\
		
		PromptIR~\cite{36}
		& \cellcolor[HTML]{DFE8FE}{34.34} & \cellcolor[HTML]{ffc7ce}{0.938}
		& \cellcolor[HTML]{DFE8FE}{31.71} & \cellcolor[HTML]{ffc7ce}{0.897}
		& \cellcolor[HTML]{DFE8FE}{28.49} & \cellcolor[HTML]{DFE8FE}{0.813}
		& 34.77 & \cellcolor[HTML]{ffc7ce}{0.952}
		& 32.49 & 0.929
		& 29.39 & 0.881 \\
		
		DFPIR~\cite{42}
		& 34.32 & 0.934
		& \cellcolor[HTML]{DFE8FE}{31.71} & \cellcolor[HTML]{ffc7ce}{0.897}
		& \cellcolor[HTML]{DFE8FE}{28.49} & \cellcolor[HTML]{ffc7ce}{0.814}
		& \cellcolor[HTML]{DFE8FE}{34.79} & \cellcolor[HTML]{ffc7ce}{0.952}
		& \cellcolor[HTML]{DFE8FE}{32.57} & \cellcolor[HTML]{DFE8FE}{0.930}
		& \cellcolor[HTML]{DFE8FE}{29.53} & \cellcolor[HTML]{DFE8FE}{0.883} \\
		
		\midrule
		
		QuReC (Ours)
		& \cellcolor[HTML]{ffc7ce}{34.41} & \cellcolor[HTML]{DFE8FE}{0.936}
		& \cellcolor[HTML]{ffc7ce}{31.78} & \cellcolor[HTML]{DFE8FE}{0.895}
		& \cellcolor[HTML]{ffc7ce}{28.56} & \cellcolor[HTML]{DFE8FE}{0.813}
		& \cellcolor[HTML]{ffc7ce}{35.08} & \cellcolor[HTML]{ffc7ce}{0.952}
		& \cellcolor[HTML]{ffc7ce}{32.91} & \cellcolor[HTML]{ffc7ce}{0.931}
		& \cellcolor[HTML]{ffc7ce}{29.93} & \cellcolor[HTML]{ffc7ce}{0.888} \\
		
		\bottomrule
	\end{tabular}
\end{table*}
\subsection{Comparison with Dedicated Single-Task Methods}
Table~\ref{tab_sup_8} presents the quantitative comparison on the dehazing task. Although QuReC is designed as a unified all-in-one restoration framework, it still achieves the best performance among both classical dehazing methods and recent restoration models. Specifically, QuReC obtains 32.13 dB in PSNR and 0.983 in SSIM, outperforming the previous best result achieved by DFPIR. This result suggests that the proposed framework not only generalizes well across diverse degradation types, but also remains highly competitive even in a dedicated single-task setting. Table~\ref{tab_sup_9} reports the comparison on the deraining task. QuReC achieves the best performance with 39.14 dB PSNR and 0.986 SSIM, slightly but consistently surpassing DFPIR, PromptIR, Restormer, and other representative deraining methods. This result further verifies that the proposed query-specific guidance is beneficial even when the degradation type is relatively consistent, since it still helps the model perform more targeted restoration and preserve fine image details. Table~\ref{tab_sup_10} shows the denoising results on CBSD68 and Urban100 under three Gaussian noise levels. QuReC achieves the best PSNR in all six evaluation settings, demonstrating strong denoising capability on both natural-image and urban-scene benchmarks. On CBSD68, QuReC reaches 34.41 dB, 31.78 dB, and 28.56 dB for noise levels 15, 25, and 50, respectively. On Urban100, it further achieves 35.08 dB, 32.91 dB, and 29.93 dB, all of which are the best results among the compared methods. In terms of SSIM, QuReC also achieves highly competitive performance, obtaining the best or tied-best results in multiple settings. These results further confirm that, despite being designed for unified restoration, QuReC does not sacrifice task-specific performance, but instead delivers consistently strong results across different denoising benchmarks.

\subsection{Additional Qualitative Comparisons under the All-in-One Setting with Three Degradations}
Figure~\ref{figure2}, Figure~\ref{figure3}, and Figure~\ref{figure4} provide additional qualitative comparisons for denoising, deraining, and dehazing, respectively, under the all-in-one setting with three degradations. As shown in Figure~\ref{figure2}, QuReC removes image noise more thoroughly while better preserving fine textures and edge structures, leading to cleaner and more natural restored results. Compared with other methods, it introduces fewer over-smoothing artifacts and retains richer local details. As shown in Figure~\ref{figure3}, QuReC suppresses rain streaks more effectively and restores clearer object boundaries and texture details. In contrast, some competing methods still leave visible rain residues or produce locally blurred regions. These results show that QuReC can better distinguish degradation patterns from underlying scene structures, thereby achieving more faithful deraining results. Figure~\ref{figure4} presents the dehazing comparisons. QuReC restores clearer visibility, more natural contrast, and more realistic colors, especially in distant regions and heavily degraded areas. Compared with previous methods, it produces fewer artifacts such as haze residuals, color distortion, or over-enhancement. Overall, these qualitative results are consistent with the quantitative comparisons and further verify the effectiveness of QuReC for unified restoration under diverse degradation conditions.

\section{Additional Analyses and Methodological Details}

In this section, we provide additional analyses and methodological details of QuReC, including its computational overhead, robustness to unseen degradations, the optimization procedure of weakly supervised prototype matching, and spatial visualizations of query-wise prototype responses.

\subsection{Computational Overhead Analysis}

Table~\ref{tab:computational_overhead} compares the model complexity and inference cost of QuReC with representative all-in-one image restoration methods. All latency results are measured on an NVIDIA RTX 3090 using an input resolution of $256\times256$.

\begin{table}[h]
	\centering
	\small
	\setlength{\tabcolsep}{4.5pt}
	\renewcommand{\arraystretch}{1.05}
	\caption{Computational overhead comparison. FLOPs and latency are measured using an input resolution of $256\times256$ on an NVIDIA RTX 3090. The values in parentheses indicate the additional overhead relative to the preceding configuration.}
	\label{tab:computational_overhead}
	\resizebox{\linewidth}{!}{
		\begin{tabular}{lccc}
			\toprule
			\textbf{Method} & \textbf{Params (M)} & \textbf{FLOPs (G)} & \textbf{Latency (s)} \\
			\midrule
			PromptIR & 32.96 & 158.14 & 0.0885 \\
			DFPIR    & 31.10 & 151.10 & 0.1398 \\
			\midrule
			w/o DQRM, w/o LGRCM
			& 26.13 & 140.99 & 0.0832 \\
			
			w/o DQRM, w/ LGRCM
			& 28.72 {\scriptsize(+2.59)}
			& 147.37 {\scriptsize(+6.38)}
			& 0.1109 {\scriptsize(+0.0277)} \\
			
			w/ DQRM, w/ LGRCM
			& 29.61 {\scriptsize(+0.89)}
			& 150.64 {\scriptsize(+3.27)}
			& 0.1121 {\scriptsize(+0.0012)} \\
			\bottomrule
		\end{tabular}
	}
\end{table}

As shown in Table~\ref{tab:computational_overhead}, QuReC maintains competitive model complexity and computational cost compared with representative global-hint and prompt-based restoration methods. Starting from the configuration equipped with LGRCM, introducing DQRM incurs only an additional 0.89M parameters, 3.27G FLOPs, and 0.0012s latency. The larger portion of the additional cost originates from LGRCM, which performs local-global response calibration. Overall, QuReC provides query-wise spatial degradation guidance with acceptable computational overhead.

\subsection{Robustness to Unseen Degradations}

We further evaluate the robustness of QuReC under an out-of-distribution setting involving an unseen degradation. Specifically, Gaussian noise is added to the low-light, haze, and snow composite-degradation test images, denoted as $\mathrm{L}{+}\mathrm{H}{+}\mathrm{S}$, while the prototype bank and all model parameters remain unchanged during evaluation. Gaussian noise is not included as an additional prototype in this experiment.

\begin{table}[h]
	\centering
	\small
	\setlength{\tabcolsep}{8pt}
	\renewcommand{\arraystretch}{1.05}
	\caption{SSIM comparison under unseen Gaussian noise added to the $\mathrm{L}{+}\mathrm{H}{+}\mathrm{S}$ composite-degradation test images. The prototype bank remains fixed during evaluation.}
	\label{tab:ood_evaluation}
	\begin{tabular}{lcc}
		\toprule
		\textbf{OOD setting} & \textbf{Input SSIM} & \textbf{QuReC SSIM} \\
		\midrule
		$\mathrm{L}{+}\mathrm{H}{+}\mathrm{S}{+}\mathrm{N}_{\sigma=5}$  & 0.431 & 0.682 \\
		$\mathrm{L}{+}\mathrm{H}{+}\mathrm{S}{+}\mathrm{N}_{\sigma=10}$ & 0.332 & 0.585 \\
		$\mathrm{L}{+}\mathrm{H}{+}\mathrm{S}{+}\mathrm{N}_{\sigma=15}$ & 0.259 & 0.535 \\
		$\mathrm{L}{+}\mathrm{H}{+}\mathrm{S}{+}\mathrm{N}_{\sigma=25}$ & 0.171 & 0.494 \\
		$\mathrm{L}{+}\mathrm{H}{+}\mathrm{S}{+}\mathrm{N}_{\sigma=50}$ & 0.081 & 0.351 \\
		\bottomrule
	\end{tabular}
\end{table}

As reported in Table~\ref{tab:ood_evaluation}, QuReC consistently improves SSIM over the degraded inputs across all evaluated noise levels. For example, at $\sigma=5$, QuReC improves SSIM from 0.431 to 0.682. Although restoration performance gradually decreases as the unseen noise becomes stronger, the consistent structural-quality improvements indicate that QuReC retains a certain degree of robustness when an unknown degradation is superimposed on known composite degradations. Nevertheless, severe unseen degradations remain challenging, suggesting that extending the fixed prototype space toward more flexible open-set degradation modeling is a meaningful direction for future research.

\subsection{Details of Weakly Supervised Prototype Matching}

Algorithm~\ref{alg:prototype_matching} summarizes the weakly supervised prototype matching process. Given query features and a fixed prototype bank, DQRM first computes query-wise prototype matching probabilities and reconstructs degradation-aware query guidance. Since token-level degradation annotations are unavailable, image-level degradation labels are converted into weak prototype-distribution targets. The matcher is then optimized using a weak matching loss together with mixed soft and hard prototype-usage balancing terms.

\begin{algorithm}[h]
	\caption{Weakly Supervised Prototype Matching}
	\label{alg:prototype_matching}
	\small
	\begin{algorithmic}[1]
		\REQUIRE Query features $X$, prototype bank $\mathcal{T}$, and image-level degradation labels $S_b$
		\ENSURE Prototype matching loss $\mathcal{L}_{\mathrm{pm}}$
		
		\STATE $r_{b,n}\leftarrow\phi_r(x_{b,n})$
		\STATE $\bar r_{b,n}\leftarrow r_{b,n}/\|r_{b,n}\|_2$
		\STATE $\bar t_k\leftarrow t_k/\|t_k\|_2$
		
		\STATE $\pi_{b,n,k}\leftarrow
		\dfrac{\exp(\alpha \bar r_{b,n}^{\top}\bar t_k)}
		{\sum_j\exp(\alpha \bar r_{b,n}^{\top}\bar t_j)}$
		
		\STATE Reconstruct $u_{b,n}\leftarrow\sum_k \pi_{b,n,k}t_k$
		\STATE $\bar{\pi}_{b,k}\leftarrow\frac{1}{N_b}\sum_n\pi_{b,n,k}$
		
		\STATE $y_{b,k}\leftarrow1/|S_b|$ if $k\in S_b$, and $0$ otherwise
		
		\STATE $\mathcal{L}_{\mathrm{match}}\leftarrow
		-\sum_{b,k}y_{b,k}\log(\bar{\pi}_{b,k}+\epsilon)$
		
		\STATE $u_k^{\mathrm{soft}}\leftarrow
		\frac{1}{\sum_bN_b}\sum_{b,n}\pi_{b,n,k}$
		
		\STATE $u_k^{\mathrm{hard}}\leftarrow
		\frac{1}{\sum_bN_b}
		\sum_{b,n}
		\mathbb{I}\!\left[
		k=\operatorname*{arg\,max}_{j}\pi_{b,n,j}
		\right]$
		
		\STATE $\mathcal{L}_{\mathrm{bal}}\leftarrow
		\alpha_s\mathrm{CV}^{2}(u^{\mathrm{soft}})
		+\alpha_h\mathrm{CV}^{2}(u^{\mathrm{hard}})$
		
		\STATE \textbf{return}
		$\mathcal{L}_{\mathrm{pm}}
		=\lambda_{\mathrm{bal}}\mathcal{L}_{\mathrm{bal}}
		+\lambda_{\mathrm{match}}\mathcal{L}_{\mathrm{match}}$
	\end{algorithmic}
\end{algorithm}

The weak matching term encourages the image-level average prototype distribution to remain consistent with the known degradation composition. Meanwhile, the mixed load-balancing objective regularizes prototype usage at both the probability and discrete-assignment levels. This formulation enables DQRM to learn query-wise degradation matching without requiring token-level degradation annotations and makes the matching probabilities, weak supervision targets, and prototype-usage regularization explicit.

\subsection{Visualization of Query-wise Prototype Responses}

\begin{figure}[t]
	\centering
	\includegraphics[width=1\linewidth]{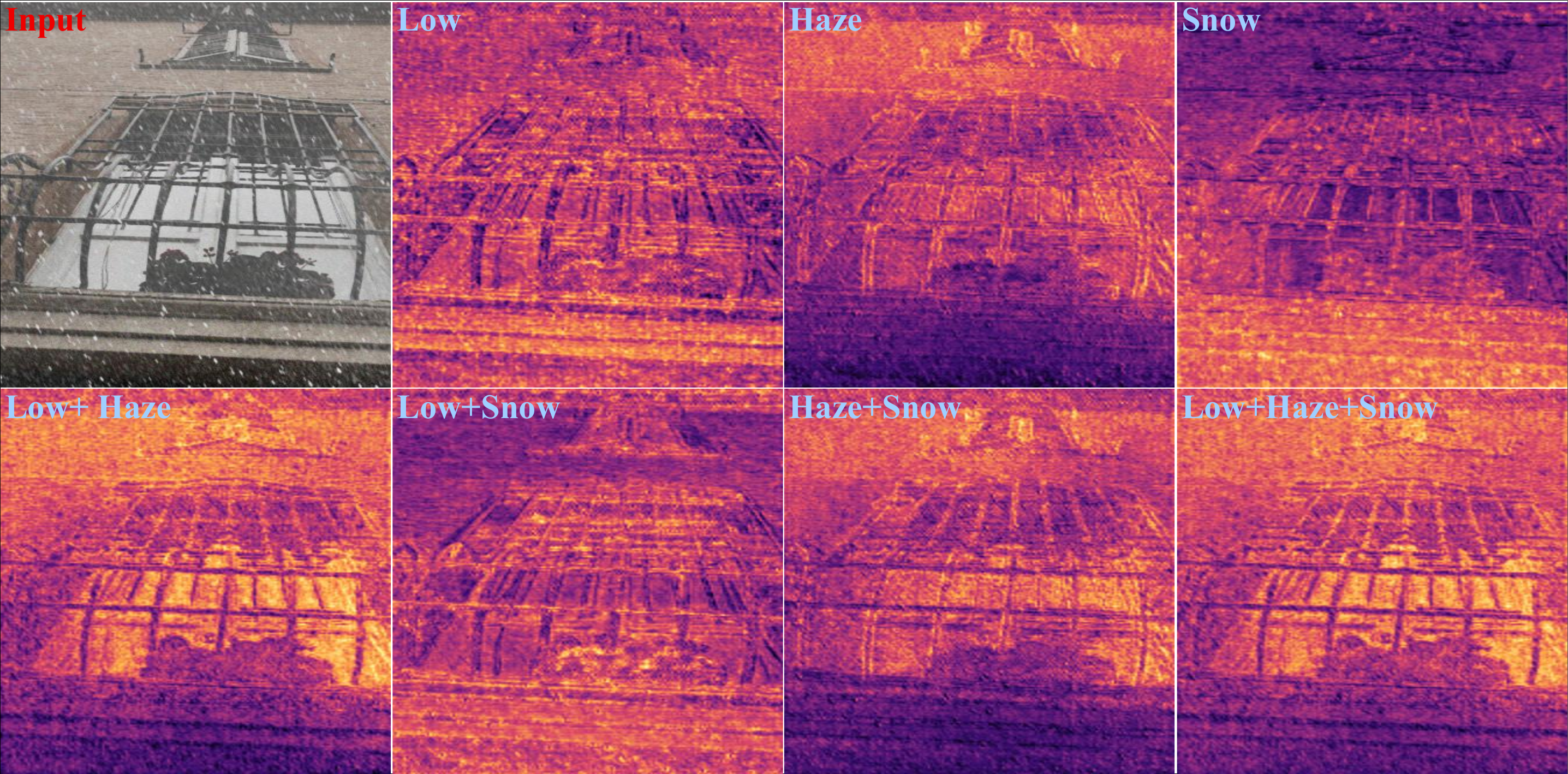}
	\caption{Spatial visualization of query-wise prototype responses on mixed-degradation images. Different degradation prototypes produce distinct spatial response patterns over the same input.}
	\label{fig:prototype_heatmaps}
\end{figure}

\begin{figure*}[t]
	\centering
	\includegraphics[width=0.98\linewidth]{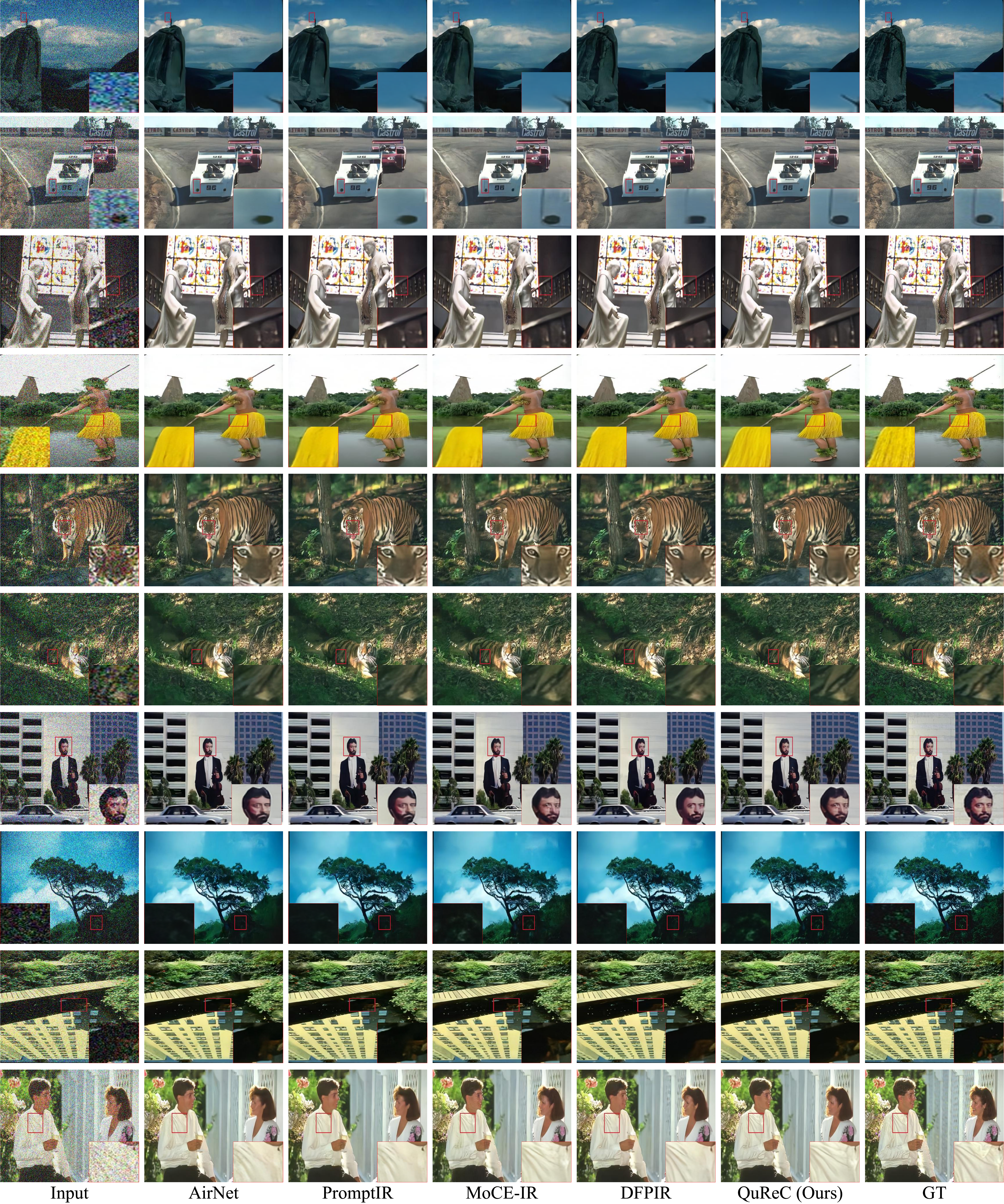}
	\caption{Additional qualitative comparisons of different methods on the denoising task under the all-in-one setting with three degradations. QuReC removes noise more effectively while better preserving fine textures and structural details.}
	\label{figure2}
\end{figure*}

\begin{figure*}[t]
	\centering
	\includegraphics[width=0.98\linewidth]{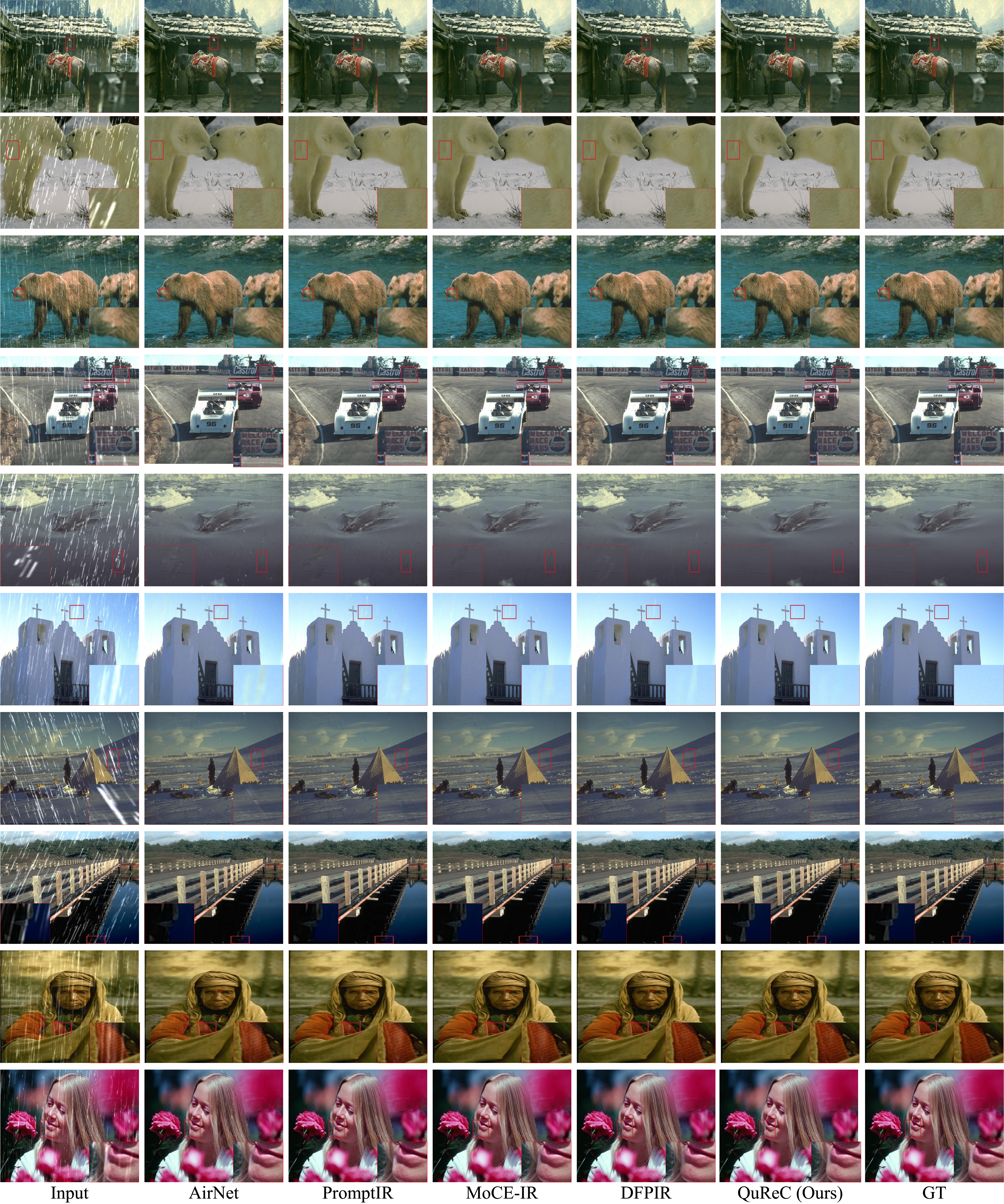}
	\caption{Additional qualitative comparisons of different methods on the deraining task under the all-in-one setting with three degradations. QuReC suppresses rain streaks more thoroughly and restores clearer structures and textures.}
	\label{figure3}
\end{figure*}

\begin{figure*}[t]
	\centering
	\includegraphics[width=0.98\linewidth]{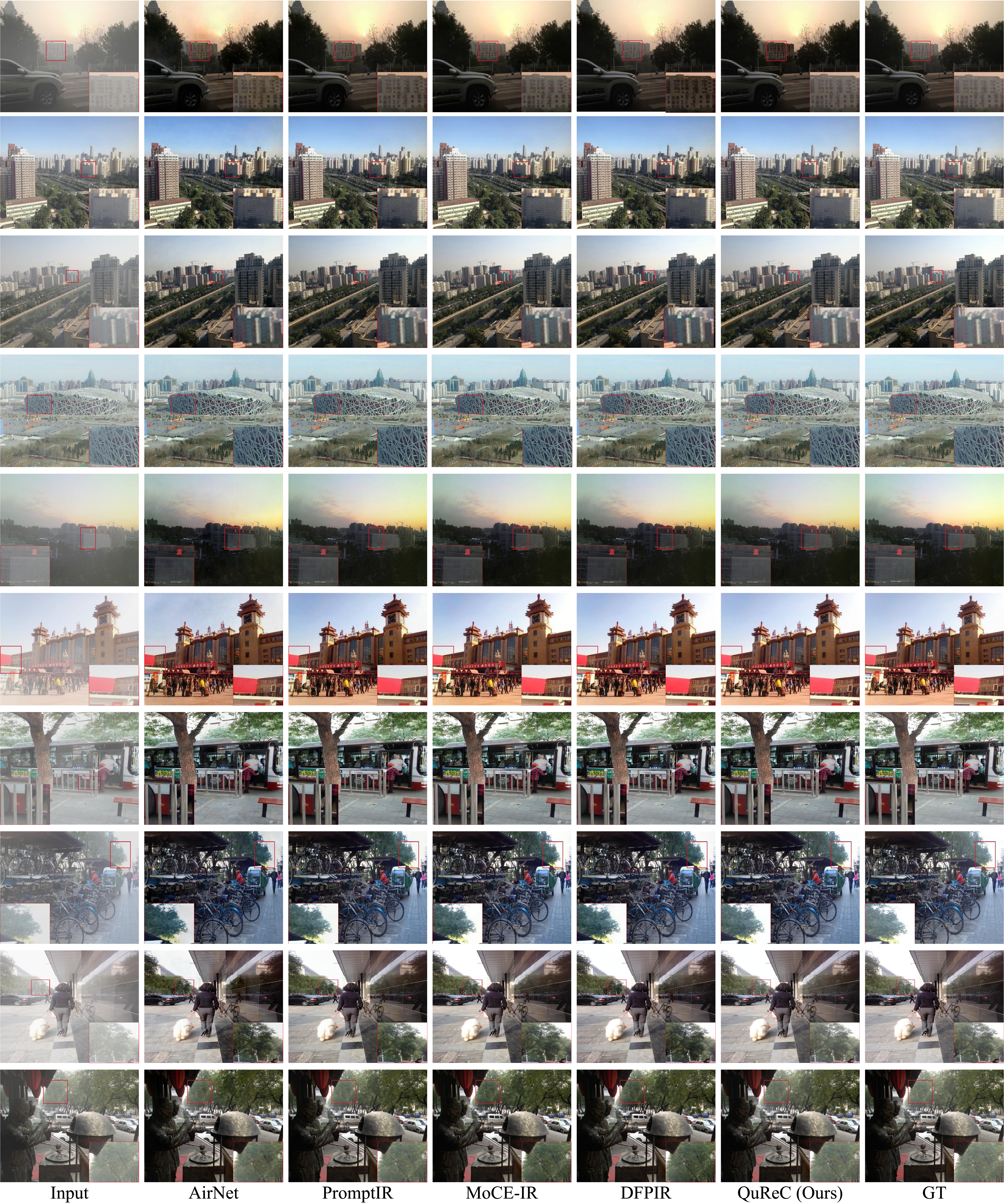}
	\caption{Additional qualitative comparisons of different methods on the dehazing task under the all-in-one setting with three degradations. QuReC restores clearer visibility, more natural contrast, and more realistic details.}
	\label{figure4}
\end{figure*}

\begin{figure*}[t]
	\centering
	\includegraphics[width=0.91\textwidth]{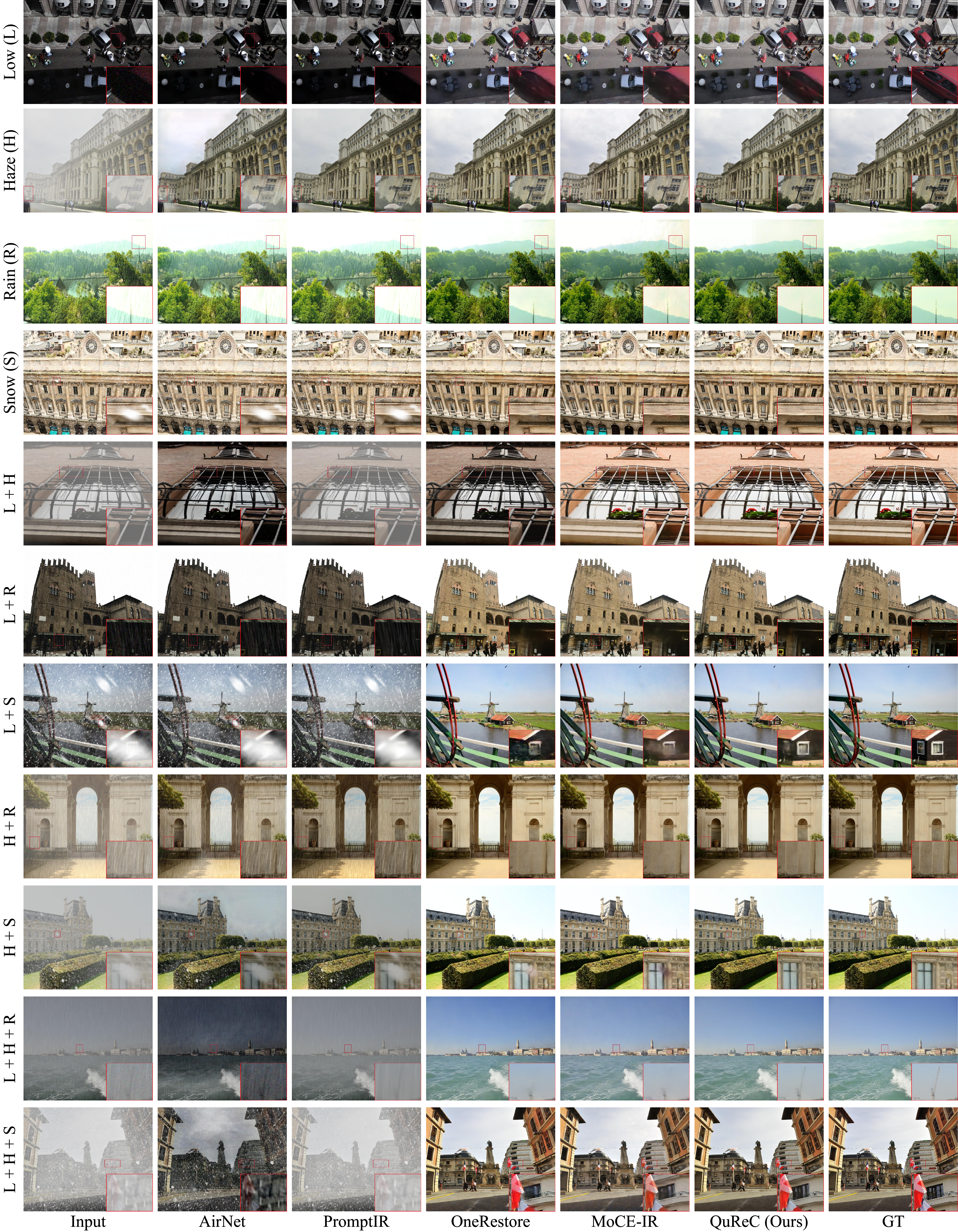}
	\caption{Qualitative comparisons of different methods across all 11 degradation cases in the CDD11 dataset. QuReC consistently restores clearer structures and more natural visual details while suppressing residual degradations more effectively under diverse composite degradation scenarios.}
	\label{figure5}
\end{figure*}

Figure~\ref{fig:prototype_heatmaps} visualizes the spatial distributions of query-wise prototype matching responses on mixed-degradation images. Different degradation prototypes exhibit distinct activation patterns over the same input image, indicating that the prototype responses are spatially non-uniform and adapt to local image content and degradation characteristics.

These visualizations provide intuitive evidence that QuReC does not rely solely on a single image-level degradation descriptor. Instead, it generates spatially varying prototype responses that provide adaptive degradation guidance for different image regions. This observation is consistent with the motivation of query-specific degradation modeling in DQRM.


\end{document}